\documentclass[11pt]{article}

\usepackage[final]{acl}

\usepackage{times}
\usepackage{latexsym}

\usepackage[T1]{fontenc}
\usepackage[utf8]{inputenc}

\usepackage{microtype}

\usepackage{inconsolata}

\usepackage{graphicx}

\usepackage{tabularx}
\usepackage{multirow}

\newcommand{\eg}{\textit{e.g.}}
\newcommand{\ie}{\textit{i.e.}}
\newcommand{\etc}{\textit{etc.}}

\usepackage[most]{tcolorbox}

\newtcolorbox{promptbox}[1][]{
  colback=gray!10,            % body background
  colframe=gray!50, % lime!50!yellow,    % border color
  coltitle=black,             % title text color
  boxrule=0.8pt,
  arc=6pt,
  left=6pt,
  right=6pt,
  top=6pt,
  bottom=6pt,
  fonttitle=\bfseries,
  title=#1,
  fontupper=\ttfamily\raggedright
}

\newcommand{\dataset}{GermanWeb}

\title{Aleph-Alpha-GermanWeb: Improving German-Language LLM Pre-Training with Model-Based Data Curation and \\ Synthetic Data Generation}

\author{%
    Thomas F Burns
    \And
    Letitia Parcalabescu
    \And
    Stephan Wäldchen
    \And
    Michael Barlow
    \AND
    Gregor Ziegltrum
    \And
    Volker Stampa
    \And
    Bastian Harren
    \And
    Björn Deiseroth\thanks{Corresponding author bjoern.deiseroth@aleph-alpha-research.com}
    \AND
    \\
    Aleph Alpha Research
}

\begin{document}
\maketitle
\begin{abstract}
Scaling data quantity is essential for large language models (LLMs), yet recent findings show that data quality can significantly boost performance and training efficiency. We introduce a German-language dataset curation pipeline that combines heuristic and model-based filtering techniques with synthetic data generation. We use our pipeline to create Aleph-Alpha-\dataset, a 628B-word German pre-training dataset composed of three subsets drawing from: (1) Common Crawl web data (organic subset; 78B words), (2) FineWeb2 (organic subset; 235B), and (3) synthetically-generated data conditioned on actual, organic web data (synthetic subset; 329B). We evaluate our dataset by pre-training both a 1B Llama-style model and an 8B tokeniser-free hierarchical autoregressive transformer (HAT) from scratch. A comparison on German-language benchmarks, including MMMLU, shows significant performance gains of Aleph-Alpha-\dataset{} over FineWeb2 alone. This advantage holds at the 8B scale even when FineWeb2 is enriched by human-curated high-quality data sources such as Wikipedia. Our findings support the growing body of evidence that model-based data curation and synthetic data generation can significantly enhance LLM pre-training datasets.
\end{abstract}

\section{Introduction} \label{sec:intro}

In recent years, ever-larger and more capable large language models (LLMs) have been released. This trend has led to the identification of power-law correlations between the loss value and the number of LLM parameters or pre-training dataset size \cite{hoffmann2022trainingcomputeoptimallargelanguage,su2024unravelingmysteryscalinglaws}. As a result, researchers  have argued \cite{hoffmann2022trainingcomputeoptimallargelanguage} that simply increasing the number of LLM parameters without also increasing the pre-training dataset size yields diminishing returns. Additionally, considering the cost of LLM inference  \cite{Siddharth2023} in production use-cases -- which can far exceed the model's initial training cost -- practitioners may weigh the cost-benefit trade-off of replacing a larger model with a smaller one to minimise inference costs \cite{Chandra2024}. To compensate for the reduced model size, a sufficiently-large training data budget would be required \cite{de2023go,hassid2024the}. However, this approach poses significant challenges for domains and languages where data is scarce.

Many recent examples in both pre-training \cite{marion2023moreinvestigatingdatapruning,NemotronCC} and post-training \cite{nguyen2025swiftcrossdatasetpruningenhancing,ye2025limoreasoning} of LLMs indicate that improvements to data quality can make training much more efficient. This is because higher-quality data can reduce the data quantity needed to reach or exceed the same model performance. In fact, conventional data scaling laws, which suggest that a small percentage increase in performance requires increasing the amount of training data by an order of magnitude, appear to be broken when data quality is improved \cite{sorscher2023neuralscalinglawsbeating}. As a result, improving data quality allows us to train LLMs much more economically. Indeed, a recent 1.7 billion parameter language model pre-trained on 11 trillion tokens achieved new state-of-the-art results for its size by focussing heavily on data quality \cite{allal2025smollm2smolgoesbig}.

Curating datasets to improve their overall quality can be broadly grouped into four categories:
\begin{enumerate}
    \item removing syntactic (near) duplication, \eg, identical documents (or with slightly different wording), which are typically straightforward to identify and remove \cite{lee2022deduplicatingtrainingdatamakes};
    \item removing semantic duplication, \eg, different summaries of the same book, which we expect to maintain some of for the purposes of syntactic variance, but should probably only keep in proportion to the level of importance or complexity of the underlying concept \cite{semdedup};
    \item removing bad data, \eg, code that does not compile \cite{CodeFuse13B} or language which is ungrammatical \cite{FineWeb}, less our models learn to repeat these mistakes; and
    \item augmenting or upsampling good data, \eg, adding high-quality synthetic data generated by LLMs \cite{long2024llmsdrivensyntheticdatageneration} or intentionally duplicating high-quality subsets of the data \cite{FineWeb2}.
\end{enumerate}

Many curation methods originated as human-led and heuristic approaches, \eg, preselecting sources with a reputation for high quality data or manual annotation and filtering of datasets \cite{RefinedWeb,FineWeb}. More recently, however, model-based approaches, \eg, training annotation models to classify data into different quality buckets \cite{DCLM} or using LLM-based text quality filtering \cite{NemotronCC}, have been shown to outperform heuristic methods. Additionally augmenting datasets with synthetically-generated data conditioned on high-quality organic data sources not only improves quantity but has been shown to further improve quality \cite{NemotronCC}.

To date, model-based and synthetic approaches exist mostly for English-language text \cite{DCLM,NemotronCC}. One of the most commonly-used and largest heuristically-filtered \textit{non}-English web datasets is FineWeb2 \cite{FineWeb2}, which adapts prior heuristic methods shown to be successful for English web data \cite{RefinedWeb,FineWeb}. Recently, model-based methods have been applied to a subset of languages from FineWeb2 to score them for quality, finding as little as 15\% of the original dataset can be used to achieve comparable performance on 1 billion parameter LLMs \cite{messmer2025enhancingmultilingualllmpretraining}.
 
Nevertheless, there remains an overall quantity issue; FineWeb2, which includes data from over 1,000 languages, is <20\% the size of the English-only FineWeb dataset \cite{FineWeb} (8TB vs. 47TB in total disk size). A single language like German constitutes <1.4\% of FineWeb2, and model-based approaches only reduce this quantity further~\cite{messmer2025enhancingmultilingualllmpretraining}. Meanwhile, synthetic approaches to improve the quantity, diversity, and quality of non-English language pre-training data have focussed on machine translation \cite{wang2025multilinguallanguagemodelpretraining}, which comes with a host of socio-technical issues \cite{Moorkens_2024}, and may lack the distinct syntactic or semantic advantages offered by synthetic approaches for LLM pre-training \cite{maini2024rephrasingwebrecipecompute}. Here we aim to fill these gaps by constructing a dataset for state-of-the-art German pre-training.

\paragraph{Contributions}

In this paper we present a data curation pipeline tailored to  German-language LLM pre-training. We use it to derive our own dataset, Aleph-Alpha-\dataset{}, a 642B-word corpus composed of three complementary subsets drawn from three distinct sources: (1) Common Crawl \cite{CommonCrawl} web data not included in FineWeb2 (78B words); (2) FineWeb2 (ODC-By v1.0) (235B words); and (3) synthetic data conditioned on actual, organic web data (329B words). Our contributions are:
\begin{enumerate}
    \item To curate the Common Crawl data, we applied a pipeline similar to (but which we show can perform better than) FineWeb2. We then augmented the dataset with synthetically-generated data and applied model-based quality classification methods;
    \item We trained two different model architectures, a 1 billion parameter Llama-style model \cite{grattafiori2024llama3herdmodels} and a tokeniser-free 8 billion parameter model \cite{neitemeier2025hierarchicalautoregressivetransformerscombining}, on each subset of \dataset{} in isolation, without mixing data across subsets. For comparison, we trained the same architectures on FineWeb2 under identical training conditions. With these models, we evaluated a range of German language benchmarks, including Multilingual Massive Multitask Language Understanding (MMMLU) \cite{hendrycks2021measuringmassivemultitasklanguage}. Our evaluations show that all three subsets of our dataset (synthetic, filtered Common Crawl, and our filtering of Fineweb2) outperform FineWeb2 \cite{FineWeb2}, even when the latter is combined with human-curated high-quality data sources such as Wikipedia at the 8 billion parameter scale. Our findings support the growing body of evidence that model-based data curation and synthetic data generation can significantly enhance pre-training datasets; and
    \item We make \dataset{} publicly available to the research community\footnote{ \url{https://huggingface.co/datasets/Aleph-Alpha/Aleph-Alpha-GermanWeb}}, in the hope that it will contribute to advancements in German-language LLM pre-training.
\end{enumerate}

\section{Methods} \label{sec:methods}

Our data curation methodology consists of three approaches: a data filtering pipeline for Common Crawl documents, synthetic document generation, and quality bucketing of all FineWeb2 documents. 

\subsection{Curating Common Crawl}\label{sec:filtering}

We followed the RefinedWeb pipeline \cite{RefinedWeb}, as adapted for German in FineWeb2 \cite{FineWeb2}, with some adjustments. We implemented all steps of the pipeline using NeMo Curator \cite{NeMoCurator} (Apache-2.0). We summarise the number of web documents our pipeline yielded in Table \ref{tab:CC-pipeline}.

\begin{table*}[t]
\centering
\begin{tabular*}{\textwidth}{@{\extracolsep{\fill}}|lrrrrrrr|}
\hline
 \textbf{CC dump} & 2024-38 & 2024-42 & 2024-46 & 2024-51 & 2025-05 & 2025-08 & \textbf{Total} \\
\hline
Original & 1,893.59 & 1,659.31 & 1,774.51 & 1,745.08 & 2,026.56 & 1,780.59 & 10,879.64 \\
URL + Lang. & 114.14 & 99.38 & 107.58 & 107.29 & 127.27 & 108.78 & 664.43 \\
Content & 65.13 & 55.90 & 60.75 & 60.29 & 72.65 & 61.20 & 375.92 \\
Exact Dedup. & 45.39 & 38.05 & 41.40 & 47.35 & 49.65 & 48.02 & 269.87 \\
Fuzzy Dedup. & 44.53 & 37.38 & 40.66 & 40.07 & 48.74 & 42.25 & 253.64 \\
\hline
Global Dedup. & & & & & & Exact & 168.19 \\
Global Dedup. & & & & & & Fuzzy & 151.61 \\
\hline
\end{tabular*}
\caption{Number of documents (in millions) throughout our curation pipeline across our downloaded Common Crawl data.}
\label{tab:CC-pipeline}
\end{table*}

\paragraph{Web data corpus.} We downloaded six Common Crawl \cite{CommonCrawl} (CC BY-SA 4.0) web data dumps, spanning from September 2024 to February 2025. Each dump captured raw web page data, e.g., HTML, metadata such as URLs, and text extracts.

\paragraph{URL filtering.} We used the URL filter developed in RefinedWeb \cite{RefinedWeb} to filter out: (i) fraudulent and adult websites from a blocklist of 4.6M domains; (ii) additional URLs based on the presence of words associated with fraudulent or adult content; and (iii) content from so-called `high-quality' websites such as Wikipedia, arXiv, \etc, which allows these datasets to be mixed in at a later point and in controlled proportions.

\paragraph{Text extraction.} Natural language texts are embedded into raw web data via HTML and other web programming languages. Since we wish to focus on natural language, we extract the natural language from the raw web data using an extraction tool. RefinedWeb used the \textsc{trafilatura} text extractor \cite{Trafilatura}. However, subsequent work \cite{DCLM} constructing an English-language dataset from Common Crawl data showed that using the \textsc{resiliparse} text extractor \cite{Resiliparse} performed better on downstream LLM evaluations. We therefore use \textsc{resiliparse} for text extraction.

\paragraph{Language identification.} RefinedWeb \cite{RefinedWeb} used the fastText language classifier of CCNet \cite{ccnet} at the document level. This classifier was trained on character n-grams from Wikipedia and supports 176 natural languages. Using this classifier, we removed documents for which the top language score was not German.

\paragraph{Repetition removal.} Web-scraped documents can include repetitive phrases or words. Heuristic methods for detection and removal of these repetitions were notably developed by \citet{Gopher}. We used these methods with the repetition identification and removal thresholds set in FineWeb2 \cite{FineWeb2} for German.

\begin{table}[htbp]
\centering

\begin{tabular}{|l|r|l|}
\hline
\textbf{Repetition Type} & \textbf{Threshold} & \textbf{Metric} \\
\hline
Duplicate lines & $28.2\%$ & \# of lines \\
% \hline
Duplicate paras & $30.0\%$ & \# of lines \\
% \hline
Duplicate paras & $20.0\%$ & \# of paras \\
% \hline
Repeated chars & $20.0\%$ & \# of lines \\
% \hline
Top 2-grams & $7.7\%$ & \# of chars \\
% \hline
Top 3-grams & $10.1\%$ & \# of chars \\
% \hline
Top 4-grams & $12.3\%$ & \# of chars \\
% \hline
Duplicate 5-grams & $14.2\%$ & \# of chars \\
% \hline
Duplicate 6-grams & $12.7\%$ & \# of chars \\
% \hline
Duplicate 7-grams & $11.5\%$ & \# of chars \\
% \hline
Duplicate 8-grams & $10.6\%$ & \# of chars \\
% \hline
Duplicate 9-grams & $9.7\%$ & \# of chars \\
% \hline
Duplicate 10-grams & $8.8\%$ & \# of chars \\
\hline
\end{tabular}
\caption{Repetition removal thresholds for methods introduced in \citet{Gopher}, with thresholds set as in FineWeb2 \cite{FineWeb2} for German. In the table, `paras' is short for `paragraphs' and `chars' is short `characters'.}\label{tab:repetitions}
\end{table}

For example, we discarded the document if it consists of more than 28.2\% duplicate lines. This example is shown in the first row of Table \ref{tab:repetitions}, where we also list the other repetition types we removed according to different thresholds for document length and document length measurement units.

\paragraph{Document-wise filtering.} We used additional document-level heuristics introduced in \citet{Gopher} and adapted in FineWeb2 \cite{FineWeb2}; namely, we required documents to have: (i) >50 and <100,000 words; (ii) a mean word length of <14 characters; (iii) a symbols-to-words ratio <0.1, where symbols are \textsc{“\#”} and \textsc{“...”}; (iv) <90\% of lines starting with a bullet point; (v) <30\% of lines ending with ellipses; (vi) >77.4\% of words containing at least one alphabetic character; and (vii) at least two of the following list of common German words: \textsc{“der”, “und”, “die”, “in”, “von”, “im”, “den”, “des”, “mit”, “das”, “er”, “dem”, “als”, “wurde”, “für”} (English translations: \textsc{“the”, “and”, “the”, “in”, “of”, “in”, “the”, “of”, “with”, “that”, “he”, “the”, “as”, “was”, “for”}).

\paragraph{Line-based filtering.} \citet{Gopher} also introduced line-wise corrections to replace or remove unwanted text within documents. We opted instead for a stricter setting by removing documents if:     (i) >15\% of the document consists of numbers; (ii) >50\% of the lines have >50\% uppercase characters; (iii) the average words per line is <10 words; or (iv) >40\% of paragraphs contain boilerplate strings, e.g., “terms of use”, “privacy policy”, \etc.

\paragraph{Exact deduplication.} By first hashing the text of each document, we performed exact deduplication to remove all but one copy of identical documents, \ie, where there are two or more documents whose strings are equal in the dataset, we kept only one of these documents. We performed this deduplication procedure over the entire dataset. This approach is notably unlike FineWeb2 \cite{FineWeb2}, which retained a certain number of duplicate documents according to their duplication rate within the overall dataset, a technique they refer to as `rehydration'.

\paragraph{Fuzzy deduplication.} Our document-level fuzzy deduplication methodology closely follows \citet{smith2022usingdeepspeedmegatrontrain}, employing a multi-stage process to identify and remove near-duplicates within the corpus. The procedure began with computing MinHash signatures \cite{MinHash} using character-based 5-grams with 23-character sequences, where the number 23 was calculated by assuming an average of 4.5 characters per word. We then used locality sensitive hashing (LSH) \cite{gionis1999similarity} to identify candidate duplicates and sort them into 14 buckets, which each had eight unique hashing functions. LSH buckets were then converted to edges for the connected components algorithm. This algorithm identified document clusters with high similarity, producing groups of near-duplicate documents that can then be removed from the corpus. We performed this deduplication procedure over the entire dataset.

\subsection{Synthetic data generation}

Inspired by similar work for English-language web data \cite{maini2024rephrasingwebrecipecompute, NemotronCC}, we performed synthetic data generation conditioned on actual -- what we also refer to as `organic' -- data from FineWeb2. We showcase the five prompt templates we used for synthetic data generation in Appendix~\ref{app:synth-gen-prompt}, designed to generate educational or basic rephrasings, summarisations, question-answer pairs, and lists of factual information. For all synthetic generation, we used the 12 billion parameter Mistral-Nemo-Instruct-2407\footnote{\url{https://huggingface.co/mistralai/Mistral-Nemo-Instruct-2407} (Apache 2.0)} LLM.

In all cases, if the entire document was longer than a specified character threshold, we segmented it into chunks of characters with lengths equal to or less than the threshold. This was done to prevent the organic document from potentially taking up all remaining context space in the prompt and to improve the quality of the LLM's responses -- prior studies \cite{maini2024rephrasingwebrecipecompute, NemotronCC} also noted for English-language that if too much organic text is inserted it degrades the synthetic data quality. To avoid splitting the documents mid-word, -sentence, or -paragraph, and thereby potentially giving impractical or nonsensical inputs to the LLM, we segmented the document texts using \textsc{text-splitter}\footnote{\url{https://github.com/benbrandt/text-splitter}}. For a given prompt and document, we synthesized data for all semantic chunks until the organic document's data was depleted. For each synthesised output, we also performed some post-processing using regular expressions and filtering to remove what we call `LLM artifacts', such as prefixes like ``Here's the rephrased version:'' and repetitions of the prompt or input text.

For a large number of FineWeb2 documents, we synthesized data based on one or more of our prompts. This resulted in many documents, especially long ones (which were segmented into more chunks), being represented in multiple ways in the final synthetic dataset. As mentioned in \citet{NemotronCC}, there is a non-negative (albeit diminishing) performance gain seen with exactly the same data during multi-epoch pre-training \cite{muennighoff2023scalingdataconstrainedlanguagemodels}, therefore it is reasonable to expect that additional `synthetic epochs' conditioned on the same underlying organic data is not harmful up to a similar limit. In the work of \citet{muennighoff2023scalingdataconstrainedlanguagemodels}, this limit was shown to be approximately four epochs. We thus chose to limit our prompt templates to five, such that there would be no more than five possible synthetic outputs conditioned on the same organic input when the whole dataset was used for training.

\subsection{Quality classification}

Following prior work on quality classification for English-language web data \cite{NemotronCC}, we categorised documents into five quality buckets -- high, medium-high, medium, medium-low, and low -- by leveraging the strengths of several imperfect quality classifiers to produce a more accurate outcome. Specifically, we created several individual quality classifiers based on fastText \cite{fasttext2016} and Bidirectional Encoder Representations from Transformers (BERT) \cite{devlin-etal-2019-bert} models. In the following, we describe (1) our two grammar classifiers, (2) our two educational quality classifiers, (3) our two instruction classifiers and (4) the way in which we ensembled all classifiers and grouped the data into the five quality buckets according to the classifiers' scores.

\subsubsection{Grammar}

We used LanguageTool\footnote{\url{https://dev.languagetool.org}} to annotate a random subset of 400,000 German FineWeb2 documents with the DE\_AGREEMENT rule, which identifies text passages with grammatical disagreement. To train our classifiers, we randomly selected 75,000 documents without identified grammar mistakes as high quality examples. As low quality examples, we took 75,000 random documents containing at least one identified grammar error.

We trained a fastText binary classifier on 95\% of the data to classify the high and low quality examples, using the remaining 5\% for validation. The model reached 63\% precision and 63\% recall on the validation set. We further trained a BERT-based binary classifier on the same train-validation splits, reaching a precision of 67\% and recall of 66\% on the validation set.

\begin{table*}[t]
\centering
\renewcommand{\arraystretch}{1.2}
\begin{tabularx}{\textwidth}{|l|X|c|}
\hline
\textbf{Model} & \textbf{Condition} & \textbf{Points} \\
\hline
-- & Initialise all documents & 0 \\
BERT (edu. quality) & Predicted score = 5 & +3 \\
% \hline
fastText (edu. quality) & Predicted “high quality” with >99\% confidence & +2 \\
% \hline
BERT (grammar) & Predicted “high quality” & +3 \\
% \hline
fastText (grammar) & Predicted “high quality” with >99\% confidence & +2 \\
BERT (instruction) & High quality probability is in top 15\% of the distribution & +6 \\
fastText (instruction) & High quality probability is in top 15\% of the distribution & +4 \\

\hline
\end{tabularx}
\caption{Overall quality scoring rules by model, condition, and assigned points.}
\label{tab:quality-points}
\end{table*}

\subsubsection{Educational quality}

Inspired by successful demonstrations for English with FineWeb \cite{FineWeb} and Nemotron-CC \cite{NemotronCC} (Apachee 2.0), we created educational quality classifiers to curate high-quality German web data by training a classifier on scores given by an LLM-as-a-judge. The intuition behind this is to identify web documents that are more likely to be educational, informative, and useful for model training, as opposed to spam, low-quality, or otherwise unhelpful content.

We first use an LLM-as-a-judge to label a small subset of the data. We then leverage these labelled documents as training data to train more efficient and lightweight fastText and BERT classifiers. This two-stage approach enables us to score a large number of web documents in a computationally-efficient way, while at the same time introducing implicit regularisation to the judging process.

We again made use of Mistral-Nemo-Instruct-2407, this time to annotate a random set of 600,000 documents from German FineWeb2 according to three criteria: (1) content quality, assessing coherence, informativeness and overall quality of the content; (2) language quality, evaluating the use of language, including formality, objectivity, and the presence of errors or slang; and (3) orthography, assessing the correctness of grammar, spelling, and punctuation, including errors such as typos, incorrect verb conjugation, and incorrect declension. We show the prompt used for the LLM-as-a-judge in Appendix \ref{app:llm-judge-prompt}, which provided scores from one to five for each criterion.

For each document, we calculated a combined \emph{educational quality score} by taking the minimum over the three criteria rated by the LLM-as-a-judge. This is because we observed that the judge would often miss poor quality in one criterium, but rarely in all three. We then used these scores as the training signal for fastText and BERT quality classification models. We trained a BERT model tasked to predict the scores given the first 512 tokens of the document's text, and a binary fastText classifier.

For the binary fastText classifier, the low- and high-quality training data subsets consisted of 185,403 documents each. The low quality subset consisted of documents with educational quality scores of one or two, whereas the high quality subset consisted of documents scoring four or five. We used 95\% of the data (and the remaining 5\% for validation) to train a fastText model to classify between high- and low-quality data. It reached 77\% precision and 77\% recall on the validation set.

To train the BERT classifier, we randomly selected a maximum of 75,000 documents from each class, which we had previously labelled with the LLM-as-a-judge with scores ranging from one to five. The resulting dataset consisted of 75,000 documents for each score above one, and 25,981 documents with scores of one. We used 95\% of this dataset for training to predict the one to five scores. The model achieved an overall accuracy of 42\% and a macro-average accuracy of 46\% when evaluated on the remaining 5\% of the data, which served as the validation set.

\subsubsection{Instruction}

Building on \citet{messmer2025enhancingmultilingualllmpretraining}, we trained binary classifiers using both fastText and BERT to distinguish between chat/instruction-style data (high quality) and other types of documents (low quality). To create a balanced training dataset, we created two subsets of 80,000 documents each. The low-quality subset consisted of randomly selected documents from the German FineWeb2 corpus, while the high-quality subset comprised randomly sampled instances from the Aya collection, Include Base-44, OpenAssistant2, and MMLU. To create samples from MMLU, we first concatenated a question and the respective answers. For OpenAssistant-2, we concatenated all samples  that belong to the same conversation.
We used 95\% of the data for training and the remaining 5\% for validation.

The fastText classifier indicated strong discriminative performance, achieving 99\% precision and 99\% recall on the validation set. Similarly, the BERT-based classifier reached a validation accuracy of 100\%.

\subsubsection{Ensembling}

To split the data into different quality buckets, we use the rules in Table \ref{tab:quality-points} to allocate \textit{overall quality} points to each document. Using these scores, we categorised documents into five quality buckets based on the overall quality, as shown in Table \ref{tab:quality_buckets}.

\begin{table}[h]
    \centering
    \begin{tabular}{|l|c|c c|}
        \hline
        \multirow{2}{*}{\textbf{Bucket}} & \multirow{2}{*}{\textbf{Score}} & 
        \multicolumn{2}{c|}{\textbf{Quality Distribution}} \\
        \cline{3-4}
        & & \textbf{Docs} & \textbf{Bytes} \\
        \hline
        High & $\geq 12$ & 12.1\% & 8.1\% \\
        Medium-high & $9-11$ & 15.0\% & 9.9\% \\
        Medium & $5-8$ & 36.3\% & 28.7\% \\
        Medium-low & $3-4$ & 15.5\% & 21.7\% \\
        Low & $< 3$ & 21.0\% & 31.6\% \\
        \hline
    \end{tabular}
    \caption{FineWeb2 document quality classifications decided by overall quality score.}
    \label{tab:quality_buckets}
\end{table}

\section{Experiments} \label{sec:experiments}

We trained and evaluated a one billion parameter (1B) Llama-style transformer model and an eight billion parameter (8B) hierarchical autoregressive transformer (HAT) model on all three subsets of our dataset in isolation to assess their quality compared to FineWeb2.

\subsection{Evaluations}

We measured model performance using the following German-language question-answering tasks.

\paragraph{MMMLU} The Multilingual Massive Multitask Language Understanding dataset~\cite{hendrycks2020measuring}, is a benchmark designed to evaluate the performance of LLMs across multiple languages and disciplines. It extends the original MMLU benchmark by translating its test set into 14 languages—including German—using professional human translators to ensure accuracy. The dataset encompasses 57 subjects ranging from elementary-level topics to advanced professional fields such as law, physics, history, and computer science. The German portion contains 14,000 samples.

The following three popular LLM benchmarks were translated from English to German~\cite{pluester_germanbenchmark}.

\paragraph{ARC} The AI2 Reasoning Challenge (ARC) \cite{allenai:arc} is a benchmark dataset developed by the Allen Institute for AI to evaluate the reasoning capabilities of artificial intelligence models. It comprises 7,787 multiple-choice science questions sourced from standardized tests designed for students in grades three through nine. The dataset is divided into two subsets: ARC-Easy and ARC-Challenge; here we evaluate the translation of ARC-Easy.

\paragraph{HellaSwag} The HellaSwag benchmark dataset \cite{zellers2019hellaswag} is designed to evaluate the commonsense reasoning abilities of AI models, particularly in the context of sentence completion tasks. It comprises approximately 70,000 multiple-choice questions from diverse sources, including instructional videos and articles from platforms like WikiHow and ActivityNet. Each question presents a context followed by four possible sentence completions, one of which is correct.

\paragraph{TruthfulQA} The TruthfulQA dataset~\cite{lin2021truthfulqa} is a benchmark designed to evaluate the truthfulness of language models when generating answers to questions. It comprises 817 questions across 38 categories, including health, law, finance, and politics. The questions are crafted to challenge models with scenarios where humans might hold incorrect beliefs or misconceptions, aiming to assess whether models can avoid generating false answers learned from imitating human texts.

\begin{figure}
    \centering
    \includegraphics[width=1\linewidth]{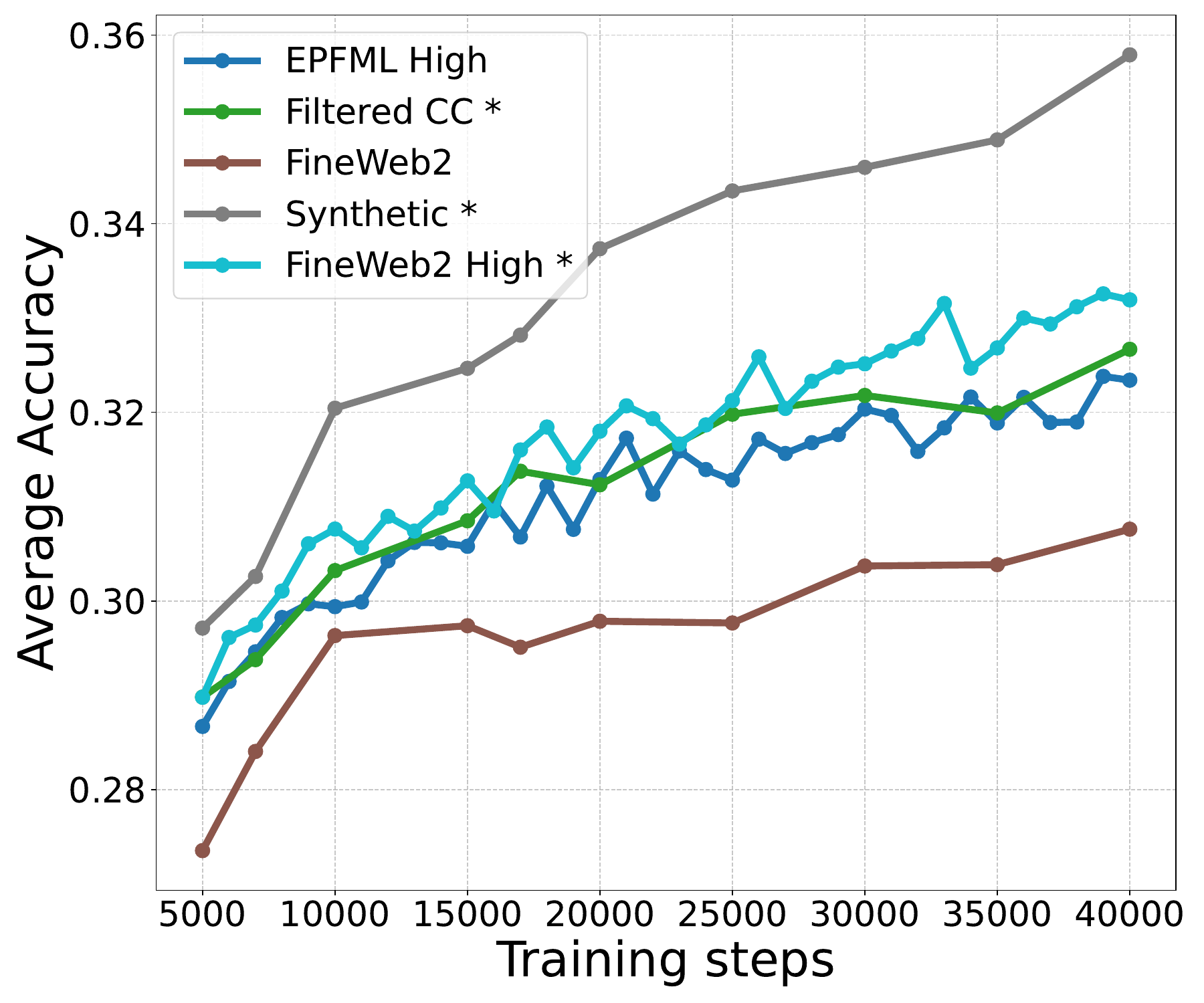}
    \caption{Average accuracies (MMMLU, ARC-Easy, HellaSwag; single- \& five-shot) of 1B Llama-style models trained on $\sim$ 84 billion tokens from different datasets. All subsets of \dataset{} -- marked with an asterisk (*) -- outperform FineWeb2. For comparison, we also include EPFML High. Here, 1,000 training steps equates to approximately 2.1 billion tokens. Individual benchmark results are provided in Appendix \ref{app:1B-benchmarks}. \label{fig:1B-overall}}
\end{figure}

\subsection{1B Llama-style model}

We pretrained 1B Llama-style models on a total budget of approximately 84 billion tokens from different datasets. Training details can be found in Appendix \ref{app:1B-training-details}.

We trained and evaluated our filtered Common Crawl (Filtered CC), synthetic, and our FineWeb2 High bucket, and compared these to random sampling German FineWeb2 and the top 10\% of German FineWeb2 documents classified by \citet{messmer2025enhancingmultilingualllmpretraining} (EPFML High). The avearge accuracies across all benchmarks are shown in Figure \ref{fig:1B-overall} and the per-benchmark breakdowns are shown in Appendix \ref{app:1B-benchmarks}. We chose to exclude TruthfulQA from the 1B parameter model evaluations, as we found that it did not give a clear signal and suspect that this size of the model may be incapable of the task (\textit{cf.} §\ref{subsec:8b-hat} for results of an 8B model).

We wish to emphasise that a large part of the synthetic subset's average performance gain stems from its far better performance on the MMLU task. Its performance on ARC is still better than FineWeb2, but not clearly the best performer. And for the Hellaswag benchmark, while it again outperforms FineWeb2, high-quality organic data appear even better. This suggests that mixing differently-derived datasets may be appropriate to balance strengths and weaknesses of individual subsets.

\subsection{8B HAT model}\label{subsec:8b-hat}

To evaluate our dataset on a different architecture and a larger scale, we pre-trained 8B hierarchical autoregressive transformer (HAT) models~\cite{neitemeierhierarchical} (see Appendix \ref{app:8B-training-details} for training details). With this we leverage recent advances demonstrating that jointly training the tokeniser and model can be a promising alternative~\cite{pagnoni2024byte}, aligning with the philosophy of end-to-end learning ~\cite{sutton2019bitter}.

An additional benefit of using an end-to-end, tokeniser-free architecture like HAT is that it debiases comparisons between datasets---comparisons which are a central focus of this paper. Tokenisers have a significant impact on the efficacy of model training~\cite{ali2024tokenizer,yang2024problematic,wang2024tokenization}. Typically they are trained on large text corpora, which makes models using those tokenisers more effective on datasets that resemble the tokeniser's training distribution. As a result, using a tokeniser-based model can introduce bias when comparing the effectiveness of different pre-training datasets, favouring those models that align more closely with the tokeniser's original training data~\cite{mayilvahanan2025llms}. To avoid this bias, we use a HAT model, which does not rely on any pre-trained tokeniser and instead performs fixed, dataset-agnostic, byte-level string splitting.

\begin{figure}[!htb]
\centering
\minipage{0.4\textwidth}
  \includegraphics[width=\linewidth]{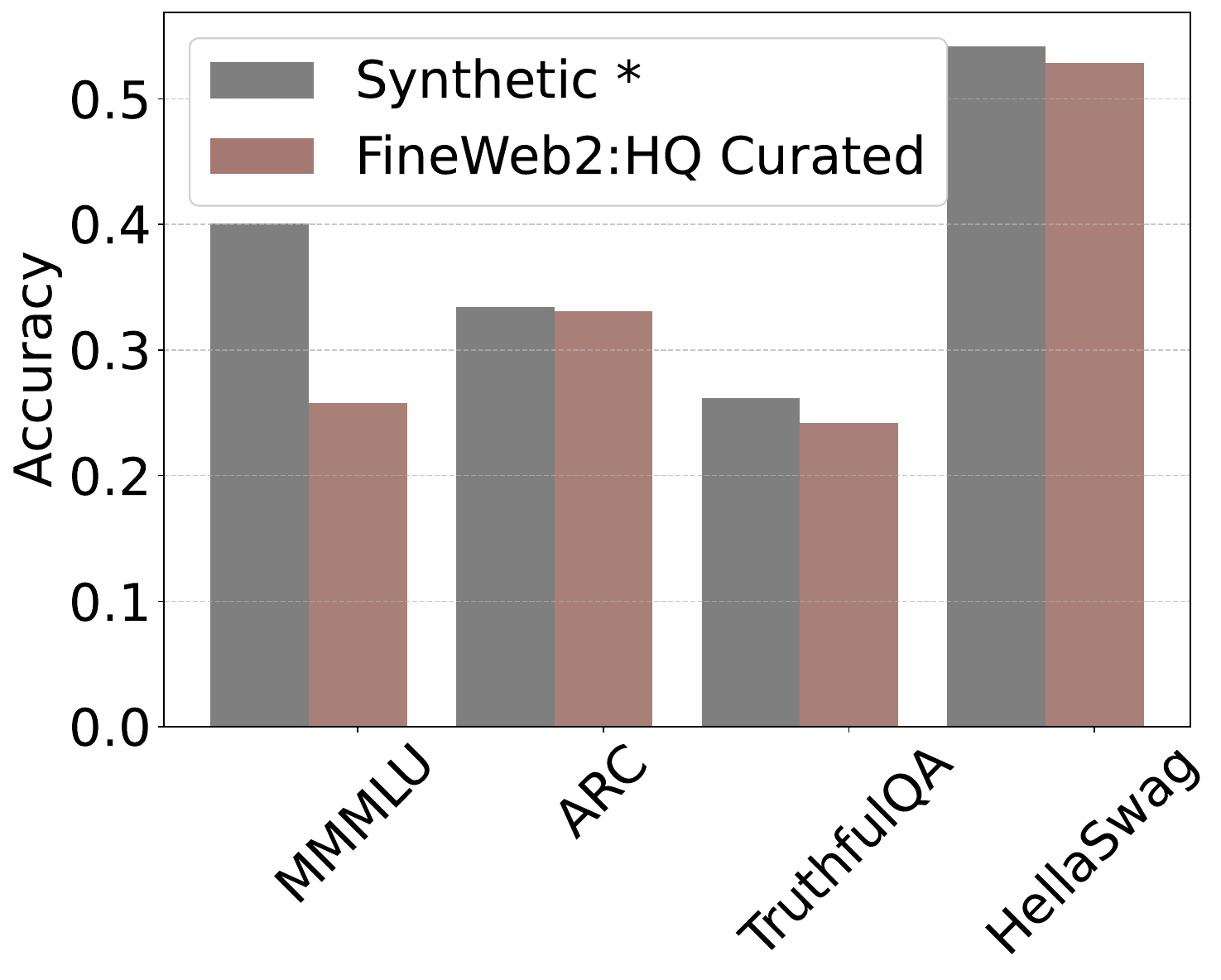}
\endminipage\hfill
\minipage{0.4\textwidth}
  \includegraphics[width=\linewidth]{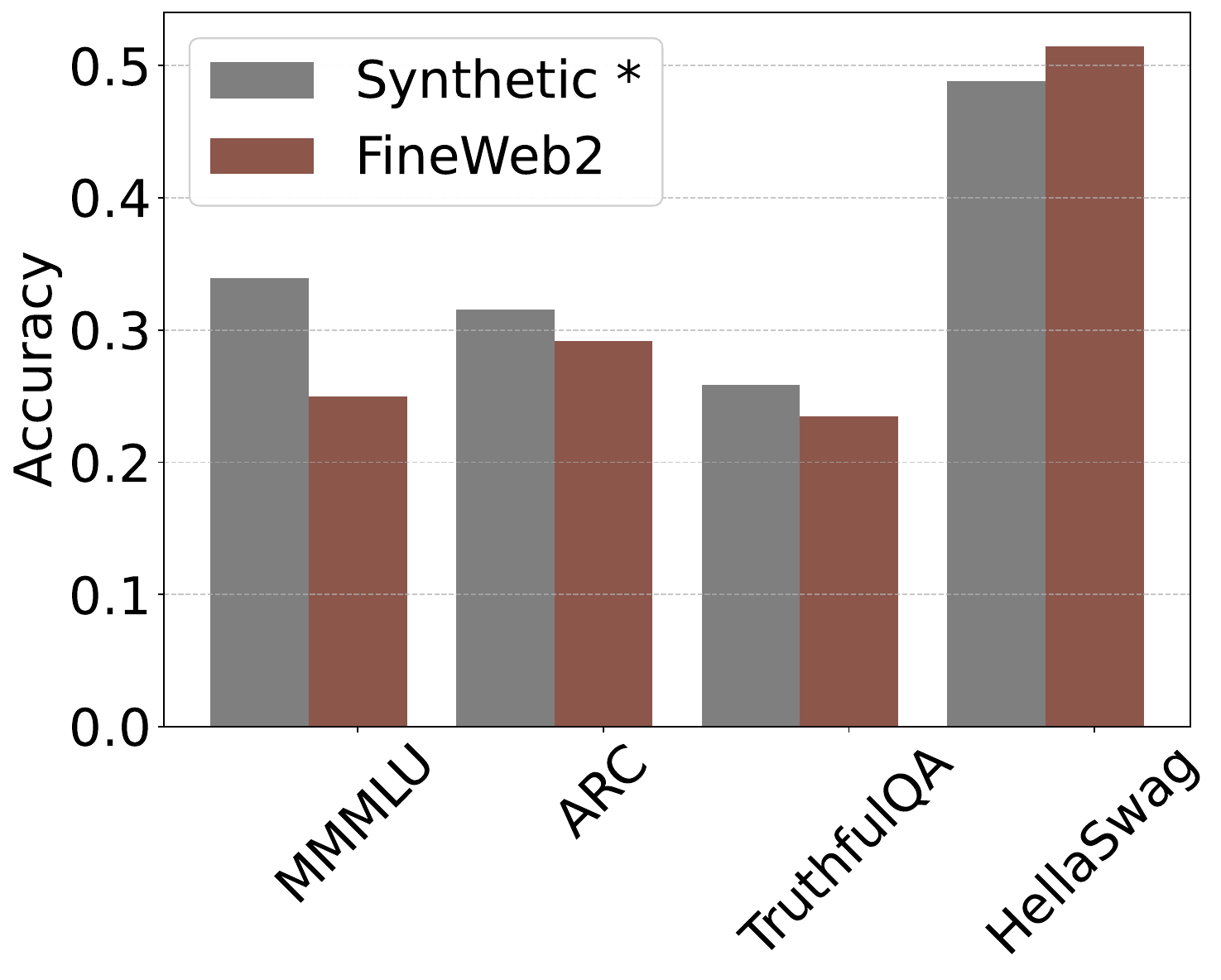}
\endminipage\hfill
\minipage{0.4\textwidth}%
  \includegraphics[width=\linewidth]{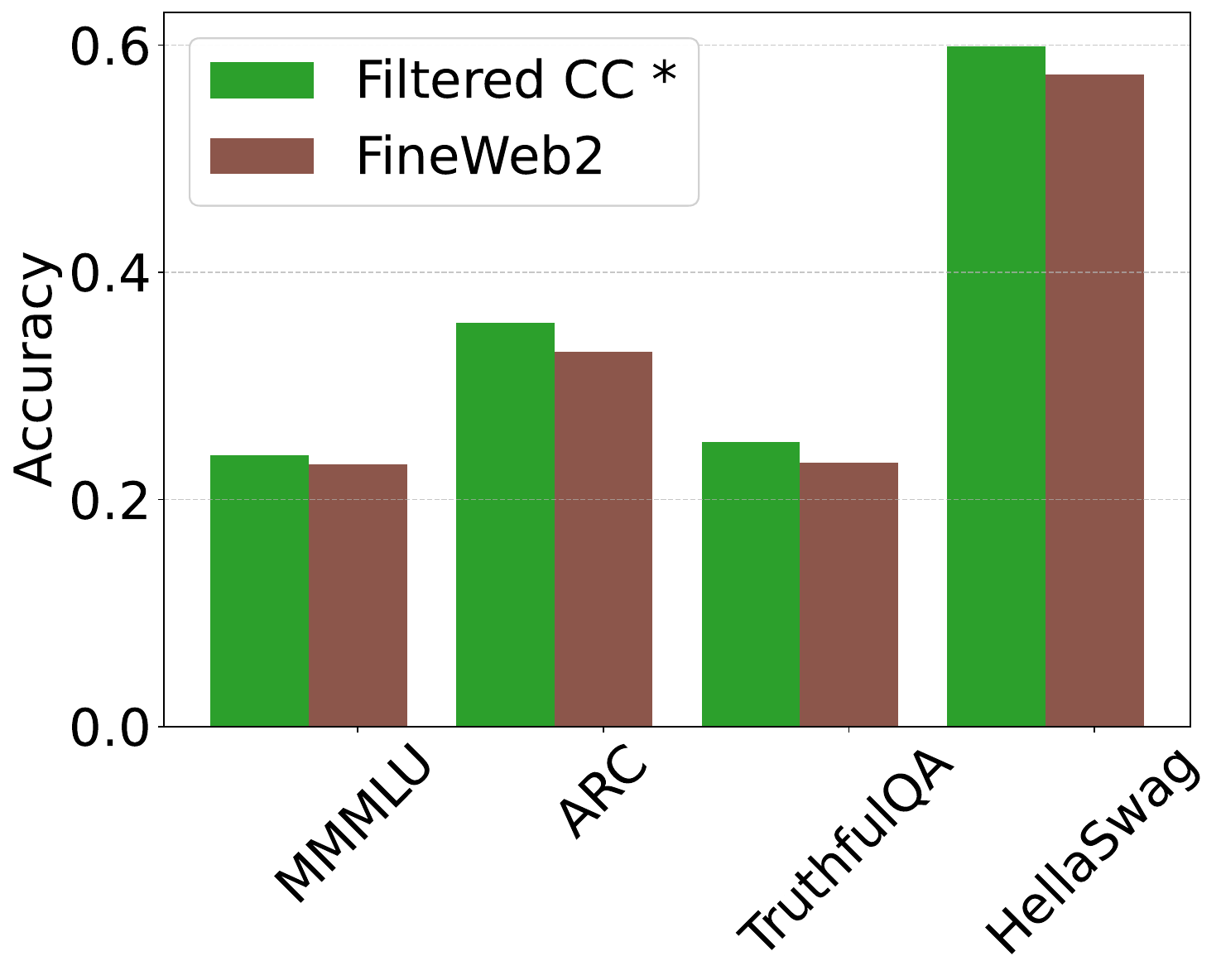}
\endminipage
\caption{Accuracies (0-shot) of 8B HAT models trained on different datasets. On average, and for all but one individual benchmark comparison, \dataset{} subsets -- marked with an asterisk (*) -- outperform FineWeb2. \textbf{(Top)} We trained for 25,000 training steps on an English-language web-derived dataset, equating to $\sim$ 75 billion English words. Afterwards the model was further trained for 20,000 steps ($\sim$~60 billion words) with German language data, either random FineWeb2 data augmented with high quality datasets like Wikipedia (FineWeb2:HQ Curated), and listed further in Appendix \ref{app:HQ-datasets-list}, or our generated synthetic data (Synthetic). Both datasets amount to 60 billion German words each.
\textbf{(Middle)} We trained for 21,000 steps ($\sim$~63 billion words) on either random FineWeb2 data or on our synthetic German data. \textbf{(Bottom)} The FineWeb2 training run was continued to 50,0000 steps and is compared to a training of 50,000 steps on data from our Filtered CC pipeline. Both runs each equate to $\sim$~150 billion words.
}\label{fig:8B-results}
\end{figure}

Figure \ref{fig:8B-results} compares the performance of these models pre-trained for different datasets and training scenarios. Differently to the 1B Llama-style models, these 8B HAT models operate on words, not tokens, resulting in 1,000 training steps equating to approximately three billion words in this set-up. We confirmed that our synthetic data on average outperforms randomly-sampled FineWeb2 data in Figure \ref{fig:8B-results} (centre). We also confirmed our previous finding of our filtered CC data outperforming FineWeb2 at the 1B scale in Figure \ref{fig:8B-results} (right). 

When pre-training LLMs, web data is often mixed from different languages and with data from alternative, curated sources \cite{grattafiori2024llama3herdmodels}, \eg, English and German data might be mixed with additional non-web data from books and encyclopedias. To test the performance of our single highest-performing subset, we compared our synthetic data to mix of 50\% FineWeb2 and 50\% high-quality curated datasets\footnote{In particular: the German National Library, European Union Parliament transcripts, European Patent Office applications, German Text Archive, Project Gutenberg books, JRC-Acquis, NewsAPI, Open Legal Data, Open Web Math, peS2o, PhilPapers, Wikibooks, Wikinews, Wikpedia, Wikisource, and Wikivoyage.}. Figure \ref{fig:8B-results} (left) shows our synthetic data outperformed this mixture on our tested benchmarks.

\section{Conclusion}\label{sec:discussion-conclusion}

This paper presents a comprehensive data curation pipeline for German-language LLM pre-training, resulting in the creation of our high-quality German dataset, \dataset{} composed of three subsets drawn from Common Crawl, FineWeb2, and synthetic data conditioned on organic web documents.  We evaluate each subset in isolation and benchmark them against FineWeb2 training demonstrating the effectiveness of our data curation pipeline and synthetic data generation in enhancing pre-training datasets. We trained two different model architectures on \dataset{} and FineWeb2 on a range of German-language benchmarks. Our results show that all three subsets of our dataset individually outperform FineWeb2, even when the latter is combined with high-quality human-curated data sources. This highlights the potential of future work (i) iterating on existing data curation pipelines, (ii) improving synthetic data generation, and (iii) mixing organic and synthetic data for advancing German-language LLM pre-training.

We make \dataset{} publicly available to the research community to contribute to further advancements in this field. Our findings support the importance of data curation and synthetic data generation in LLM pre-training, and we believe that our work will have a positive impact on the development of German-language LLMs.

\section{Limitations}

We believe there are many improvements and further research directions which will be important in this area. For example, we expect high-quality translation of existing, high-performing English datasets -- such as explored in \citet{wang2025multilinguallanguagemodelpretraining} -- to be a helpful augmentation method. In our manual inspection of machine-translated texts from English to German, however, we found a high number of errors. We therefore caution against training on large amounts of machine-translated data without also rigorously assessing its quality and naturalness.

Relatedly, a more thorough analysis of synthetic data and its apparent quality is needed. Recent work has demonstrated that recursively training models exclusively on synthetic data leads to a phenomenon termed `model collapse' \cite{shumailov2024ai}, wherein model performance gradually deteriorates over successive generations of training and synthesis. However, mixing synthetic data with organic data appears to avoid this issue \cite{gerstgrasser2024modelcollapseinevitablebreaking}. Still, it remains unclear to what degree practitioners should opt for organic rather than synthetic data, or whether synthetic data is merely `gaming' popular evaluation methods by providing data in a more relevant syntax than organic data.

\bibliography{custom}

@misc{RefinedWeb,
      title={The RefinedWeb Dataset for Falcon LLM: Outperforming Curated Corpora with Web Data, and Web Data Only}, 
      author={Guilherme Penedo and Quentin Malartic and Daniel Hesslow and Ruxandra Cojocaru and Alessandro Cappelli and Hamza Alobeidli and Baptiste Pannier and Ebtesam Almazrouei and Julien Launay},
      year={2023},
      eprint={2306.01116},
      archivePrefix={arXiv},
      primaryClass={cs.CL},
      url={https://arxiv.org/abs/2306.01116}, 
}

@InProceedings{Resiliparse,
  address =             {Berlin Heidelberg New York},
  author =              {Janek Bevendorff and Benno Stein and Matthias Hagen and Martin Potthast},
  booktitle =           {Advances in Information Retrieval. 40th European Conference on IR Research (ECIR 2018)},
  editor =              {Leif Azzopardi and Allan Hanbury and Gabriella Pasi and Benjamin Piwowarski},
  ids =                 {potthast:2018c,stein:2018c},
  month =               mar,
  publisher =           {Springer},
  series =              {Lecture Notes in Computer Science},
  site =                {Grenoble, France},
  title =               {{Elastic ChatNoir: Search Engine for the ClueWeb and the Common Crawl}},
  year =                2018
}

@misc{muennighoff2023scalingdataconstrainedlanguagemodels,
      title={Scaling Data-Constrained Language Models}, 
      author={Niklas Muennighoff and Alexander M. Rush and Boaz Barak and Teven Le Scao and Aleksandra Piktus and Nouamane Tazi and Sampo Pyysalo and Thomas Wolf and Colin Raffel},
      year={2023},
      eprint={2305.16264},
      archivePrefix={arXiv},
      primaryClass={cs.CL},
      url={https://arxiv.org/abs/2305.16264}, 
}

@INPROCEEDINGS{MinHash,
  author={Broder, A.Z.},
  booktitle={Proceedings. Compression and Complexity of SEQUENCES 1997 (Cat. No.97TB100171)}, 
  title={On the resemblance and containment of documents}, 
  year={1997},
  volume={},
  number={},
  pages={21-29},
  doi={10.1109/SEQUEN.1997.666900}}

@inproceedings{gionis1999similarity,
  title={Similarity search in high dimensions via hashing},
  author={Gionis, Aristides and Indyk, Piotr and Motwani, Rajeev and others},
  booktitle={Vldb},
  volume={99},
  pages={518--529},
  year={1999}
}

@misc{smith2022usingdeepspeedmegatrontrain,
      title={Using DeepSpeed and Megatron to Train Megatron-Turing NLG 530B, A Large-Scale Generative Language Model}, 
      author={Shaden Smith and Mostofa Patwary and Brandon Norick and Patrick LeGresley and Samyam Rajbhandari and Jared Casper and Zhun Liu and Shrimai Prabhumoye and George Zerveas and Vijay Korthikanti and Elton Zhang and Rewon Child and Reza Yazdani Aminabadi and Julie Bernauer and Xia Song and Mohammad Shoeybi and Yuxiong He and Michael Houston and Saurabh Tiwary and Bryan Catanzaro},
      year={2022},
      eprint={2201.11990},
      archivePrefix={arXiv},
      primaryClass={cs.CL},
      url={https://arxiv.org/abs/2201.11990}, 
}

@inproceedings{CodeFuse13B,
author = {Di, Peng and Li, Jianguo and Yu, Hang and Jiang, Wei and Cai, Wenting and Cao, Yang and Chen, Chaoyu and Chen, Dajun and Chen, Hongwei and Chen, Liang and Fan, Gang and Gong, Jie and Gong, Zi and Hu, Wen and Guo, Tingting and Lei, Zhichao and Li, Ting and Li, Zheng and Liang, Ming and Liao, Cong and Liu, Bingchang and Liu, Jiachen and Liu, Zhiwei and Lu, Shaojun and Shen, Min and Wang, Guangpei and Wang, Huan and Wang, Zhi and Xu, Zhaogui and Yang, Jiawei and Ye, Qing and Zhang, Gehao and Zhang, Yu and Zhao, Zelin and Zheng, Xunjin and Zhou, Hailian and Zhu, Lifu and Zhu, Xianying},
title = {CodeFuse-13B: A Pretrained Multi-lingual Code Large Language Model},
year = {2024},
isbn = {9798400705014},
publisher = {Association for Computing Machinery},
address = {New York, NY, USA},
url = {https://doi.org/10.1145/3639477.3639719},
doi = {10.1145/3639477.3639719},
booktitle = {Proceedings of the 46th International Conference on Software Engineering: Software Engineering in Practice},
pages = {418–429},
numpages = {12},
location = {Lisbon, Portugal},
series = {ICSE-SEIP '24}
}

@inproceedings{FineWeb,
 author = {Penedo, Guilherme and Kydl\'{\i}\v{c}ek, Hynek and allal, Loubna Ben and Lozhkov, Anton and Mitchell, Margaret and Raffel, Colin and Von Werra, Leandro and Wolf, Thomas},
 booktitle = {Advances in Neural Information Processing Systems},
 editor = {A. Globerson and L. Mackey and D. Belgrave and A. Fan and U. Paquet and J. Tomczak and C. Zhang},
 pages = {30811--30849},
 publisher = {Curran Associates, Inc.},
 title = {The FineWeb Datasets: Decanting the Web for the Finest Text Data at Scale},
 url = {https://proceedings.neurips.cc/paper_files/paper/2024/file/370df50ccfdf8bde18f8f9c2d9151bda-Paper-Datasets_and_Benchmarks_Track.pdf},
 volume = {37},
 year = {2024}
}

@misc{hendrycks2021measuringmassivemultitasklanguage,
      title={Measuring Massive Multitask Language Understanding}, 
      author={Dan Hendrycks and Collin Burns and Steven Basart and Andy Zou and Mantas Mazeika and Dawn Song and Jacob Steinhardt},
      year={2021},
      eprint={2009.03300},
      archivePrefix={arXiv},
      primaryClass={cs.CY},
      url={https://arxiv.org/abs/2009.03300}, 
}

@misc{grattafiori2024llama3herdmodels,
      title={The Llama 3 Herd of Models}, 
      author={Aaron Grattafiori and Abhimanyu Dubey and Abhinav Jauhri and Abhinav Pandey and Abhishek Kadian and Ahmad Al-Dahle and Aiesha Letman and Akhil Mathur and Alan Schelten and Alex Vaughan and Amy Yang and Angela Fan and Anirudh Goyal and Anthony Hartshorn and Aobo Yang and Archi Mitra and Archie Sravankumar and Artem Korenev and Arthur Hinsvark and Arun Rao and Aston Zhang and Aurelien Rodriguez and Austen Gregerson and Ava Spataru and Baptiste Roziere and Bethany Biron and Binh Tang and Bobbie Chern and Charlotte Caucheteux and Chaya Nayak and Chloe Bi and Chris Marra and Chris McConnell and Christian Keller and Christophe Touret and Chunyang Wu and Corinne Wong and Cristian Canton Ferrer and Cyrus Nikolaidis and Damien Allonsius and Daniel Song and Danielle Pintz and Danny Livshits and Danny Wyatt and David Esiobu and Dhruv Choudhary and Dhruv Mahajan and Diego Garcia-Olano and Diego Perino and Dieuwke Hupkes and Egor Lakomkin and Ehab AlBadawy and Elina Lobanova and Emily Dinan and Eric Michael Smith and Filip Radenovic and Francisco Guzmán and Frank Zhang and Gabriel Synnaeve and Gabrielle Lee and Georgia Lewis Anderson and Govind Thattai and Graeme Nail and Gregoire Mialon and Guan Pang and Guillem Cucurell and Hailey Nguyen and Hannah Korevaar and Hu Xu and Hugo Touvron and Iliyan Zarov and Imanol Arrieta Ibarra and Isabel Kloumann and Ishan Misra and Ivan Evtimov and Jack Zhang and Jade Copet and Jaewon Lee and Jan Geffert and Jana Vranes and Jason Park and Jay Mahadeokar and Jeet Shah and Jelmer van der Linde and Jennifer Billock and Jenny Hong and Jenya Lee and Jeremy Fu and Jianfeng Chi and Jianyu Huang and Jiawen Liu and Jie Wang and Jiecao Yu and Joanna Bitton and Joe Spisak and Jongsoo Park and Joseph Rocca and Joshua Johnstun and Joshua Saxe and Junteng Jia and Kalyan Vasuden Alwala and Karthik Prasad and Kartikeya Upasani and Kate Plawiak and Ke Li and Kenneth Heafield and Kevin Stone and Khalid El-Arini and Krithika Iyer and Kshitiz Malik and Kuenley Chiu and Kunal Bhalla and Kushal Lakhotia and Lauren Rantala-Yeary and Laurens van der Maaten and Lawrence Chen and Liang Tan and Liz Jenkins and Louis Martin and Lovish Madaan and Lubo Malo and Lukas Blecher and Lukas Landzaat and Luke de Oliveira and Madeline Muzzi and Mahesh Pasupuleti and Mannat Singh and Manohar Paluri and Marcin Kardas and Maria Tsimpoukelli and Mathew Oldham and Mathieu Rita and Maya Pavlova and Melanie Kambadur and Mike Lewis and Min Si and Mitesh Kumar Singh and Mona Hassan and Naman Goyal and Narjes Torabi and Nikolay Bashlykov and Nikolay Bogoychev and Niladri Chatterji and Ning Zhang and Olivier Duchenne and Onur Çelebi and Patrick Alrassy and Pengchuan Zhang and Pengwei Li and Petar Vasic and Peter Weng and Prajjwal Bhargava and Pratik Dubal and Praveen Krishnan and Punit Singh Koura and Puxin Xu and Qing He and Qingxiao Dong and Ragavan Srinivasan and Raj Ganapathy and Ramon Calderer and Ricardo Silveira Cabral and Robert Stojnic and Roberta Raileanu and Rohan Maheswari and Rohit Girdhar and Rohit Patel and Romain Sauvestre and Ronnie Polidoro and Roshan Sumbaly and Ross Taylor and Ruan Silva and Rui Hou and Rui Wang and Saghar Hosseini and Sahana Chennabasappa and Sanjay Singh and Sean Bell and Seohyun Sonia Kim and Sergey Edunov and Shaoliang Nie and Sharan Narang and Sharath Raparthy and Sheng Shen and Shengye Wan and Shruti Bhosale and Shun Zhang and Simon Vandenhende and Soumya Batra and Spencer Whitman and Sten Sootla and Stephane Collot and Suchin Gururangan and Sydney Borodinsky and Tamar Herman and Tara Fowler and Tarek Sheasha and Thomas Georgiou and Thomas Scialom and Tobias Speckbacher and Todor Mihaylov and Tong Xiao and Ujjwal Karn and Vedanuj Goswami and Vibhor Gupta and Vignesh Ramanathan and Viktor Kerkez and Vincent Gonguet and Virginie Do and Vish Vogeti and Vítor Albiero and Vladan Petrovic and Weiwei Chu and Wenhan Xiong and Wenyin Fu and Whitney Meers and Xavier Martinet and Xiaodong Wang and Xiaofang Wang and Xiaoqing Ellen Tan and Xide Xia and Xinfeng Xie and Xuchao Jia and Xuewei Wang and Yaelle Goldschlag and Yashesh Gaur and Yasmine Babaei and Yi Wen and Yiwen Song and Yuchen Zhang and Yue Li and Yuning Mao and Zacharie Delpierre Coudert and Zheng Yan and Zhengxing Chen and Zoe Papakipos and Aaditya Singh and Aayushi Srivastava and Abha Jain and Adam Kelsey and Adam Shajnfeld and Adithya Gangidi and Adolfo Victoria and Ahuva Goldstand and Ajay Menon and Ajay Sharma and Alex Boesenberg and Alexei Baevski and Allie Feinstein and Amanda Kallet and Amit Sangani and Amos Teo and Anam Yunus and Andrei Lupu and Andres Alvarado and Andrew Caples and Andrew Gu and Andrew Ho and Andrew Poulton and Andrew Ryan and Ankit Ramchandani and Annie Dong and Annie Franco and Anuj Goyal and Aparajita Saraf and Arkabandhu Chowdhury and Ashley Gabriel and Ashwin Bharambe and Assaf Eisenman and Azadeh Yazdan and Beau James and Ben Maurer and Benjamin Leonhardi and Bernie Huang and Beth Loyd and Beto De Paola and Bhargavi Paranjape and Bing Liu and Bo Wu and Boyu Ni and Braden Hancock and Bram Wasti and Brandon Spence and Brani Stojkovic and Brian Gamido and Britt Montalvo and Carl Parker and Carly Burton and Catalina Mejia and Ce Liu and Changhan Wang and Changkyu Kim and Chao Zhou and Chester Hu and Ching-Hsiang Chu and Chris Cai and Chris Tindal and Christoph Feichtenhofer and Cynthia Gao and Damon Civin and Dana Beaty and Daniel Kreymer and Daniel Li and David Adkins and David Xu and Davide Testuggine and Delia David and Devi Parikh and Diana Liskovich and Didem Foss and Dingkang Wang and Duc Le and Dustin Holland and Edward Dowling and Eissa Jamil and Elaine Montgomery and Eleonora Presani and Emily Hahn and Emily Wood and Eric-Tuan Le and Erik Brinkman and Esteban Arcaute and Evan Dunbar and Evan Smothers and Fei Sun and Felix Kreuk and Feng Tian and Filippos Kokkinos and Firat Ozgenel and Francesco Caggioni and Frank Kanayet and Frank Seide and Gabriela Medina Florez and Gabriella Schwarz and Gada Badeer and Georgia Swee and Gil Halpern and Grant Herman and Grigory Sizov and Guangyi and Zhang and Guna Lakshminarayanan and Hakan Inan and Hamid Shojanazeri and Han Zou and Hannah Wang and Hanwen Zha and Haroun Habeeb and Harrison Rudolph and Helen Suk and Henry Aspegren and Hunter Goldman and Hongyuan Zhan and Ibrahim Damlaj and Igor Molybog and Igor Tufanov and Ilias Leontiadis and Irina-Elena Veliche and Itai Gat and Jake Weissman and James Geboski and James Kohli and Janice Lam and Japhet Asher and Jean-Baptiste Gaya and Jeff Marcus and Jeff Tang and Jennifer Chan and Jenny Zhen and Jeremy Reizenstein and Jeremy Teboul and Jessica Zhong and Jian Jin and Jingyi Yang and Joe Cummings and Jon Carvill and Jon Shepard and Jonathan McPhie and Jonathan Torres and Josh Ginsburg and Junjie Wang and Kai Wu and Kam Hou U and Karan Saxena and Kartikay Khandelwal and Katayoun Zand and Kathy Matosich and Kaushik Veeraraghavan and Kelly Michelena and Keqian Li and Kiran Jagadeesh and Kun Huang and Kunal Chawla and Kyle Huang and Lailin Chen and Lakshya Garg and Lavender A and Leandro Silva and Lee Bell and Lei Zhang and Liangpeng Guo and Licheng Yu and Liron Moshkovich and Luca Wehrstedt and Madian Khabsa and Manav Avalani and Manish Bhatt and Martynas Mankus and Matan Hasson and Matthew Lennie and Matthias Reso and Maxim Groshev and Maxim Naumov and Maya Lathi and Meghan Keneally and Miao Liu and Michael L. Seltzer and Michal Valko and Michelle Restrepo and Mihir Patel and Mik Vyatskov and Mikayel Samvelyan and Mike Clark and Mike Macey and Mike Wang and Miquel Jubert Hermoso and Mo Metanat and Mohammad Rastegari and Munish Bansal and Nandhini Santhanam and Natascha Parks and Natasha White and Navyata Bawa and Nayan Singhal and Nick Egebo and Nicolas Usunier and Nikhil Mehta and Nikolay Pavlovich Laptev and Ning Dong and Norman Cheng and Oleg Chernoguz and Olivia Hart and Omkar Salpekar and Ozlem Kalinli and Parkin Kent and Parth Parekh and Paul Saab and Pavan Balaji and Pedro Rittner and Philip Bontrager and Pierre Roux and Piotr Dollar and Polina Zvyagina and Prashant Ratanchandani and Pritish Yuvraj and Qian Liang and Rachad Alao and Rachel Rodriguez and Rafi Ayub and Raghotham Murthy and Raghu Nayani and Rahul Mitra and Rangaprabhu Parthasarathy and Raymond Li and Rebekkah Hogan and Robin Battey and Rocky Wang and Russ Howes and Ruty Rinott and Sachin Mehta and Sachin Siby and Sai Jayesh Bondu and Samyak Datta and Sara Chugh and Sara Hunt and Sargun Dhillon and Sasha Sidorov and Satadru Pan and Saurabh Mahajan and Saurabh Verma and Seiji Yamamoto and Sharadh Ramaswamy and Shaun Lindsay and Shaun Lindsay and Sheng Feng and Shenghao Lin and Shengxin Cindy Zha and Shishir Patil and Shiva Shankar and Shuqiang Zhang and Shuqiang Zhang and Sinong Wang and Sneha Agarwal and Soji Sajuyigbe and Soumith Chintala and Stephanie Max and Stephen Chen and Steve Kehoe and Steve Satterfield and Sudarshan Govindaprasad and Sumit Gupta and Summer Deng and Sungmin Cho and Sunny Virk and Suraj Subramanian and Sy Choudhury and Sydney Goldman and Tal Remez and Tamar Glaser and Tamara Best and Thilo Koehler and Thomas Robinson and Tianhe Li and Tianjun Zhang and Tim Matthews and Timothy Chou and Tzook Shaked and Varun Vontimitta and Victoria Ajayi and Victoria Montanez and Vijai Mohan and Vinay Satish Kumar and Vishal Mangla and Vlad Ionescu and Vlad Poenaru and Vlad Tiberiu Mihailescu and Vladimir Ivanov and Wei Li and Wenchen Wang and Wenwen Jiang and Wes Bouaziz and Will Constable and Xiaocheng Tang and Xiaojian Wu and Xiaolan Wang and Xilun Wu and Xinbo Gao and Yaniv Kleinman and Yanjun Chen and Ye Hu and Ye Jia and Ye Qi and Yenda Li and Yilin Zhang and Ying Zhang and Yossi Adi and Youngjin Nam and Yu and Wang and Yu Zhao and Yuchen Hao and Yundi Qian and Yunlu Li and Yuzi He and Zach Rait and Zachary DeVito and Zef Rosnbrick and Zhaoduo Wen and Zhenyu Yang and Zhiwei Zhao and Zhiyu Ma},
      year={2024},
      eprint={2407.21783},
      archivePrefix={arXiv},
      primaryClass={cs.AI},
      url={https://arxiv.org/abs/2407.21783}, 
}

@misc{neitemeier2025hierarchicalautoregressivetransformerscombining,
      title={Hierarchical Autoregressive Transformers: Combining Byte- and Word-Level Processing for Robust, Adaptable Language Models}, 
      author={Pit Neitemeier and Björn Deiseroth and Constantin Eichenberg and Lukas Balles},
      year={2025},
      eprint={2501.10322},
      archivePrefix={arXiv},
      primaryClass={cs.CL},
      url={https://arxiv.org/abs/2501.10322}, 
}

@misc{long2024llmsdrivensyntheticdatageneration,
      title={On LLMs-Driven Synthetic Data Generation, Curation, and Evaluation: A Survey}, 
      author={Lin Long and Rui Wang and Ruixuan Xiao and Junbo Zhao and Xiao Ding and Gang Chen and Haobo Wang},
      year={2024},
      eprint={2406.15126},
      archivePrefix={arXiv},
      primaryClass={cs.CL},
      url={https://arxiv.org/abs/2406.15126}, 
}

@misc{messmer2025enhancingmultilingualllmpretraining,
      title={Enhancing Multilingual LLM Pretraining with Model-Based Data Selection}, 
      author={Bettina Messmer and Vinko Sabolčec and Martin Jaggi},
      year={2025},
      eprint={2502.10361},
      archivePrefix={arXiv},
      primaryClass={cs.CL},
      url={https://arxiv.org/abs/2502.10361}, 
}

@misc{semdedup,
      title={SemDeDup: Data-efficient learning at web-scale through semantic deduplication}, 
      author={Amro Abbas and Kushal Tirumala and Dániel Simig and Surya Ganguli and Ari S. Morcos},
      year={2023},
      eprint={2303.09540},
      archivePrefix={arXiv},
      primaryClass={cs.LG},
      url={https://arxiv.org/abs/2303.09540}, 
}

@misc{lee2022deduplicatingtrainingdatamakes,
      title={Deduplicating Training Data Makes Language Models Better}, 
      author={Katherine Lee and Daphne Ippolito and Andrew Nystrom and Chiyuan Zhang and Douglas Eck and Chris Callison-Burch and Nicholas Carlini},
      year={2022},
      eprint={2107.06499},
      archivePrefix={arXiv},
      primaryClass={cs.CL},
      url={https://arxiv.org/abs/2107.06499}, 
}

@misc{ye2025limoreasoning,
      title={LIMO: Less is More for Reasoning}, 
      author={Yixin Ye and Zhen Huang and Yang Xiao and Ethan Chern and Shijie Xia and Pengfei Liu},
      year={2025},
      eprint={2502.03387},
      archivePrefix={arXiv},
      primaryClass={cs.CL},
      url={https://arxiv.org/abs/2502.03387}, 
}

@misc{nguyen2025swiftcrossdatasetpruningenhancing,
      title={Swift Cross-Dataset Pruning: Enhancing Fine-Tuning Efficiency in Natural Language Understanding}, 
      author={Binh-Nguyen Nguyen and Yang He},
      year={2025},
      eprint={2501.02432},
      archivePrefix={arXiv},
      primaryClass={cs.CL},
      url={https://arxiv.org/abs/2501.02432}, 
}

@misc{marion2023moreinvestigatingdatapruning,
      title={When Less is More: Investigating Data Pruning for Pretraining LLMs at Scale}, 
      author={Max Marion and Ahmet Üstün and Luiza Pozzobon and Alex Wang and Marzieh Fadaee and Sara Hooker},
      year={2023},
      eprint={2309.04564},
      archivePrefix={arXiv},
      primaryClass={cs.CL},
      url={https://arxiv.org/abs/2309.04564}, 
}

@misc{sorscher2023neuralscalinglawsbeating,
      title={Beyond neural scaling laws: beating power law scaling via data pruning}, 
      author={Ben Sorscher and Robert Geirhos and Shashank Shekhar and Surya Ganguli and Ari S. Morcos},
      year={2023},
      eprint={2206.14486},
      archivePrefix={arXiv},
      primaryClass={cs.LG},
      url={https://arxiv.org/abs/2206.14486}, 
}

@misc{maini2024rephrasingwebrecipecompute,
      title={Rephrasing the Web: A Recipe for Compute and Data-Efficient Language Modeling}, 
      author={Pratyush Maini and Skyler Seto and He Bai and David Grangier and Yizhe Zhang and Navdeep Jaitly},
      year={2024},
      eprint={2401.16380},
      archivePrefix={arXiv},
      primaryClass={cs.CL},
      url={https://arxiv.org/abs/2401.16380}, 
}

@misc{allal2025smollm2smolgoesbig,
      title={SmolLM2: When Smol Goes Big -- Data-Centric Training of a Small Language Model}, 
      author={Loubna Ben Allal and Anton Lozhkov and Elie Bakouch and Gabriel Martín Blázquez and Guilherme Penedo and Lewis Tunstall and Andrés Marafioti and Hynek Kydlíček and Agustín Piqueres Lajarín and Vaibhav Srivastav and Joshua Lochner and Caleb Fahlgren and Xuan-Son Nguyen and Clémentine Fourrier and Ben Burtenshaw and Hugo Larcher and Haojun Zhao and Cyril Zakka and Mathieu Morlon and Colin Raffel and Leandro von Werra and Thomas Wolf},
      year={2025},
      eprint={2502.02737},
      archivePrefix={arXiv},
      primaryClass={cs.CL},
      url={https://arxiv.org/abs/2502.02737}, 
}

@inproceedings{
hassid2024the,
title={The Larger the Better? Improved {LLM} Code-Generation via Budget Reallocation},
author={Michael Hassid and Tal Remez and Jonas Gehring and Roy Schwartz and Yossi Adi},
booktitle={First Conference on Language Modeling},
year={2024},
url={https://openreview.net/forum?id=QJvfpWSpWm}
}

@INPROCEEDINGS{Chandra2024,
  author={Irugalbandara, Chandra and Mahendra, Ashish and Daynauth, Roland and Arachchige, Tharuka Kasthuri and Dantanarayana, Jayanaka and Flautner, Krisztian and Tang, Lingjia and Kang, Yiping and Mars, Jason},
  booktitle={2024 IEEE International Symposium on Performance Analysis of Systems and Software (ISPASS)}, 
  title={Scaling Down to Scale Up: A Cost-Benefit Analysis of Replacing OpenAI's LLM with Open Source SLMs in Production}, 
  year={2024},
  volume={},
  number={},
  pages={280-291},
  doi={10.1109/ISPASS61541.2024.00034}}

@INPROCEEDINGS{Siddharth2023,
  author={Samsi, Siddharth and Zhao, Dan and McDonald, Joseph and Li, Baolin and Michaleas, Adam and Jones, Michael and Bergeron, William and Kepner, Jeremy and Tiwari, Devesh and Gadepally, Vijay},
  booktitle={2023 IEEE High Performance Extreme Computing Conference (HPEC)}, 
  title={From Words to Watts: Benchmarking the Energy Costs of Large Language Model Inference}, 
  year={2023},
  volume={},
  number={},
  pages={1-9},
  doi={10.1109/HPEC58863.2023.10363447}}

@misc{hoffmann2022trainingcomputeoptimallargelanguage,
      title={Training Compute-Optimal Large Language Models}, 
      author={Jordan Hoffmann and Sebastian Borgeaud and Arthur Mensch and Elena Buchatskaya and Trevor Cai and Eliza Rutherford and Diego de Las Casas and Lisa Anne Hendricks and Johannes Welbl and Aidan Clark and Tom Hennigan and Eric Noland and Katie Millican and George van den Driessche and Bogdan Damoc and Aurelia Guy and Simon Osindero and Karen Simonyan and Erich Elsen and Jack W. Rae and Oriol Vinyals and Laurent Sifre},
      year={2022},
      eprint={2203.15556},
      archivePrefix={arXiv},
      primaryClass={cs.CL},
      url={https://arxiv.org/abs/2203.15556}, 
}

@misc{su2024unravelingmysteryscalinglaws,
      title={Unraveling the Mystery of Scaling Laws: Part I}, 
      author={Hui Su and Zhi Tian and Xiaoyu Shen and Xunliang Cai},
      year={2024},
      eprint={2403.06563},
      archivePrefix={arXiv},
      primaryClass={cs.LG},
      url={https://arxiv.org/abs/2403.06563}, 
}

@misc{NeMoCurator,
author = {Jennings, Joseph and Patwary, Mostofa and Subramanian, Sandeep and Prabhumoye, Shrimai and Dattagupta, Ayush and Jawa, Vibhu and Liu, Jiwei and Wolf, Ryan and Yurick, Sarah and Singh, Varun},
title = {{NeMo-Curator: a toolkit for data curation}},
url = {https://github.com/NVIDIA/NeMo-Curator}
}

@misc{Gopher,
      title={Scaling Language Models: Methods, Analysis \& Insights from Training Gopher}, 
      author={Jack W. Rae and Sebastian Borgeaud and Trevor Cai and Katie Millican and Jordan Hoffmann and Francis Song and John Aslanides and Sarah Henderson and Roman Ring and Susannah Young and Eliza Rutherford and Tom Hennigan and Jacob Menick and Albin Cassirer and Richard Powell and George van den Driessche and Lisa Anne Hendricks and Maribeth Rauh and Po-Sen Huang and Amelia Glaese and Johannes Welbl and Sumanth Dathathri and Saffron Huang and Jonathan Uesato and John Mellor and Irina Higgins and Antonia Creswell and Nat McAleese and Amy Wu and Erich Elsen and Siddhant Jayakumar and Elena Buchatskaya and David Budden and Esme Sutherland and Karen Simonyan and Michela Paganini and Laurent Sifre and Lena Martens and Xiang Lorraine Li and Adhiguna Kuncoro and Aida Nematzadeh and Elena Gribovskaya and Domenic Donato and Angeliki Lazaridou and Arthur Mensch and Jean-Baptiste Lespiau and Maria Tsimpoukelli and Nikolai Grigorev and Doug Fritz and Thibault Sottiaux and Mantas Pajarskas and Toby Pohlen and Zhitao Gong and Daniel Toyama and Cyprien de Masson d'Autume and Yujia Li and Tayfun Terzi and Vladimir Mikulik and Igor Babuschkin and Aidan Clark and Diego de Las Casas and Aurelia Guy and Chris Jones and James Bradbury and Matthew Johnson and Blake Hechtman and Laura Weidinger and Iason Gabriel and William Isaac and Ed Lockhart and Simon Osindero and Laura Rimell and Chris Dyer and Oriol Vinyals and Kareem Ayoub and Jeff Stanway and Lorrayne Bennett and Demis Hassabis and Koray Kavukcuoglu and Geoffrey Irving},
      year={2022},
      eprint={2112.11446},
      archivePrefix={arXiv},
      primaryClass={cs.CL},
      url={https://arxiv.org/abs/2112.11446}, 
}

@inproceedings{ccnet,
    title = "{CCN}et: Extracting High Quality Monolingual Datasets from Web Crawl Data",
    author = "Wenzek, Guillaume  and
      Lachaux, Marie-Anne  and
      Conneau, Alexis  and
      Chaudhary, Vishrav  and
      Guzm{\'a}n, Francisco  and
      Joulin, Armand  and
      Grave, Edouard",
    editor = "Calzolari, Nicoletta  and
      B{\'e}chet, Fr{\'e}d{\'e}ric  and
      Blache, Philippe  and
      Choukri, Khalid  and
      Cieri, Christopher  and
      Declerck, Thierry  and
      Goggi, Sara  and
      Isahara, Hitoshi  and
      Maegaard, Bente  and
      Mariani, Joseph  and
      Mazo, H{\'e}l{\`e}ne  and
      Moreno, Asuncion  and
      Odijk, Jan  and
      Piperidis, Stelios",
    booktitle = "Proceedings of the Twelfth Language Resources and Evaluation Conference",
    month = may,
    year = "2020",
    address = "Marseille, France",
    publisher = "European Language Resources Association",
    url = "https://aclanthology.org/2020.lrec-1.494/",
    pages = "4003--4012",
    language = "eng",
    ISBN = "979-10-95546-34-4"
}

@inproceedings{Trafilatura,
    title = "Trafilatura: {A} Web Scraping Library and Command-Line Tool for Text Discovery and Extraction",
    author = "Barbaresi, Adrien",
    editor = "Ji, Heng  and
      Park, Jong C.  and
      Xia, Rui",
    booktitle = "Proceedings of the 59th Annual Meeting of the Association for Computational Linguistics and the 11th International Joint Conference on Natural Language Processing: System Demonstrations",
    month = aug,
    year = "2021",
    address = "Online",
    publisher = "Association for Computational Linguistics",
    url = "https://aclanthology.org/2021.acl-demo.15/",
    doi = "10.18653/v1/2021.acl-demo.15",
    pages = "122--131"
}

@misc{DCLM,
      title={DataComp-LM: In search of the next generation of training sets for language models}, 
      author={Jeffrey Li and Alex Fang and Georgios Smyrnis and Maor Ivgi and Matt Jordan and Samir Gadre and Hritik Bansal and Etash Guha and Sedrick Keh and Kushal Arora and Saurabh Garg and Rui Xin and Niklas Muennighoff and Reinhard Heckel and Jean Mercat and Mayee Chen and Suchin Gururangan and Mitchell Wortsman and Alon Albalak and Yonatan Bitton and Marianna Nezhurina and Amro Abbas and Cheng-Yu Hsieh and Dhruba Ghosh and Josh Gardner and Maciej Kilian and Hanlin Zhang and Rulin Shao and Sarah Pratt and Sunny Sanyal and Gabriel Ilharco and Giannis Daras and Kalyani Marathe and Aaron Gokaslan and Jieyu Zhang and Khyathi Chandu and Thao Nguyen and Igor Vasiljevic and Sham Kakade and Shuran Song and Sujay Sanghavi and Fartash Faghri and Sewoong Oh and Luke Zettlemoyer and Kyle Lo and Alaaeldin El-Nouby and Hadi Pouransari and Alexander Toshev and Stephanie Wang and Dirk Groeneveld and Luca Soldaini and Pang Wei Koh and Jenia Jitsev and Thomas Kollar and Alexandros G. Dimakis and Yair Carmon and Achal Dave and Ludwig Schmidt and Vaishaal Shankar},
      year={2024},
      eprint={2406.11794},
      archivePrefix={arXiv},
      primaryClass={cs.LG},
      url={https://arxiv.org/abs/2406.11794}, 
}

@Misc{CommonCrawl,
  author       = "Common Crawl Foundation",
  title        = "Common Crawl",
  URL          = "https://commoncrawl.org/",
  cc-author-affiliation = "Common Crawl",
}

@misc{NemotronCC,
      title={Nemotron-CC: Transforming Common Crawl into a Refined Long-Horizon Pretraining Dataset}, 
      author={Dan Su and Kezhi Kong and Ying Lin and Joseph Jennings and Brandon Norick and Markus Kliegl and Mostofa Patwary and Mohammad Shoeybi and Bryan Catanzaro},
      year={2024},
      eprint={2412.02595},
      archivePrefix={arXiv},
      primaryClass={cs.CL},
      url={https://arxiv.org/abs/2412.02595}, 
}

@inproceedings{
FineWeb2,
title={FineWeb2: One Pipeline to Scale Them All {\textemdash} Adapting Pre-Training Data Processing to Every Language},
author={Guilherme Penedo and Hynek Kydl{\'\i}{\v{c}}ek and Vinko Sabol{\v{c}}ec and Bettina Messmer and Negar Foroutan and Amir Hossein Kargaran and Colin Raffel and Martin Jaggi and Leandro Von Werra and Thomas Wolf},
booktitle={Second Conference on Language Modeling},
year={2025},
url={https://openreview.net/forum?id=jnRBe6zatP}
}

@article{fasttext2016,
  title={Enriching Word Vectors with Subword Information},
  author={Bojanowski, Piotr and Grave, Edouard and Joulin, Armand and Mikolov, Tomas},
  journal={arXiv preprint arXiv:1607.04606},
  year={2016}
}

@inproceedings{devlin-etal-2019-bert,
    title = "{BERT}: Pre-training of Deep Bidirectional Transformers for Language Understanding",
    author = "Devlin, Jacob  and
      Chang, Ming-Wei  and
      Lee, Kenton  and
      Toutanova, Kristina",
    editor = "Burstein, Jill  and
      Doran, Christy  and
      Solorio, Thamar",
    booktitle = "Proceedings of the 2019 Conference of the North {A}merican Chapter of the Association for Computational Linguistics: Human Language Technologies, Volume 1 (Long and Short Papers)",
    month = jun,
    year = "2019",
    address = "Minneapolis, Minnesota",
    publisher = "Association for Computational Linguistics",
    url = "https://aclanthology.org/N19-1423/",
    doi = "10.18653/v1/N19-1423",
    pages = "4171--4186"
}

@article{de2023go,
  title={Go smol or go home, 2023},
  author={De Vries, Harm},
  journal={URL https://www. harmdevries. com/post/model-size-vs-compute-overhead},
  year={2023}
}

@article{mayilvahanan2025llms,
  title={LLMs on the Line: Data Determines Loss-to-Loss Scaling Laws},
  author={Mayilvahanan, Prasanna and Wiedemer, Thadd{\"a}us and Mallick, Sayak and Bethge, Matthias and Brendel, Wieland},
  journal={arXiv preprint arXiv:2502.12120},
  year={2025}
}

@inproceedings{ali2024tokenizer,
  title={Tokenizer choice for llm training: Negligible or crucial?},
  author={Ali, Mehdi and Fromm, Michael and Thellmann, Klaudia and Rutmann, Richard and L{\"u}bbering, Max and Leveling, Johannes and Klug, Katrin and Ebert, Jan and Doll, Niclas and Buschhoff, Jasper and others},
  booktitle={Findings of the Association for Computational Linguistics: NAACL 2024},
  pages={3907--3924},
  year={2024}
}

@inproceedings{yang2024problematic,
  title={Problematic Tokens: Tokenizer Bias in Large Language Models},
  author={Yang, Jin and Wang, Zhiqiang and Lin, Yanbin and Zhao, Zunduo},
  booktitle={2024 IEEE International Conference on Big Data (BigData)},
  pages={6387--6393},
  year={2024},
  organization={IEEE}
}

@article{wang2024tokenization,
  title={Tokenization matters! degrading large language models through challenging their tokenization},
  author={Wang, Dixuan and Li, Yanda and Jiang, Junyuan and Ding, Zepeng and Jiang, Guochao and Liang, Jiaqing and Yang, Deqing},
  journal={arXiv preprint arXiv:2405.17067},
  year={2024}
}

@article{pagnoni2024byte,
  title={Byte latent transformer: Patches scale better than tokens},
  author={Pagnoni, Artidoro and Pasunuru, Ram and Rodriguez, Pedro and Nguyen, John and Muller, Benjamin and Li, Margaret and Zhou, Chunting and Yu, Lili and Weston, Jason and Zettlemoyer, Luke and others},
  journal={arXiv preprint arXiv:2412.09871},
  year={2024}
}

@article{hendrycks2020measuring,
  title={Measuring massive multitask language understanding},
  author={Hendrycks, Dan and Burns, Collin and Basart, Steven and Zou, Andy and Mazeika, Mantas and Song, Dawn and Steinhardt, Jacob},
  journal={arXiv preprint arXiv:2009.03300},
  year={2020}
}

@misc{pluester_germanbenchmark,
  author       = {Björn Plüster},
  title        = {GermanBenchmark: Translating popular LLM benchmarks to German},
  howpublished = {\url{https://github.com/bjoernpl/GermanBenchmark}},
  note         = {Accessed: 2025-04-17}
}

@article{zellers2019hellaswag,
  title={Hellaswag: Can a machine really finish your sentence?},
  author={Zellers, Rowan and Holtzman, Ari and Bisk, Yonatan and Farhadi, Ali and Choi, Yejin},
  journal={arXiv preprint arXiv:1905.07830},
  year={2019}
}

@article{allenai:arc,
      author    = {Peter Clark  and Isaac Cowhey and Oren Etzioni and Tushar Khot and
                    Ashish Sabharwal and Carissa Schoenick and Oyvind Tafjord},
      title     = {Think you have Solved Question Answering? Try ARC, the AI2 Reasoning Challenge},
      journal   = {arXiv:1803.05457v1},
      year      = {2018},
}

@misc{lin2021truthfulqa,
    title={TruthfulQA: Measuring How Models Mimic Human Falsehoods},
    author={Stephanie Lin and Jacob Hilton and Owain Evans},
    year={2021},
    eprint={2109.07958},
    archivePrefix={arXiv},
    primaryClass={cs.CL}
}

@misc{alephalpha_pharia1_2024,
  author       = {Aleph Alpha},
  title        = {Pharia-1-LLM-7B-control},
  year         = {2024},
  publisher    = {Hugging Face},
  howpublished = {\url{https://huggingface.co/Aleph-Alpha/Pharia-1-LLM-7B-control}},
  note         = {Accessed: 2025-04-17}
}

@article{sutton2019bitter,
  title={The bitter lesson},
  author={Sutton, Richard},
  journal={Incomplete Ideas (blog)},
  volume={13},
  number={1},
  pages={38},
  year={2019}
}

@inproceedings{neitemeierhierarchical,
  title={Hierarchical Autoregressive Transformers for Tokenizer-Free Language Modelling},
  author={Neitemeier, Pit and Deiseroth, Bj{\"o}rn and Eichenberg, Constantin and Balles, Lukas},
  booktitle={The Thirteenth International Conference on Learning Representations},
  year={2025}
}

@misc{wang2025multilinguallanguagemodelpretraining,
      title={Multilingual Language Model Pretraining using Machine-translated Data}, 
      author={Jiayi Wang and Yao Lu and Maurice Weber and Max Ryabinin and David Adelani and Yihong Chen and Raphael Tang and Pontus Stenetorp},
      year={2025},
      eprint={2502.13252},
      archivePrefix={arXiv},
      primaryClass={cs.CL},
      url={https://arxiv.org/abs/2502.13252}, 
}

@book{Moorkens_2024,
   title={Automating Translation},
   ISBN={9781003381280},
   url={http://dx.doi.org/10.4324/9781003381280},
   DOI={10.4324/9781003381280},
   publisher={Routledge},
   author={Moorkens, Joss and Way, Andy and Lankford, Séamus},
   year={2024},
   month=jul }

@article{shumailov2024ai,
  title={AI models collapse when trained on recursively generated data},
  author={Shumailov, Ilia and Shumaylov, Zakhar and Zhao, Yiren and Papernot, Nicolas and Anderson, Ross and Gal, Yarin},
  journal={Nature},
  volume={631},
  number={8022},
  pages={755--759},
  year={2024},
  publisher={Nature Publishing Group UK London}
}

@misc{gerstgrasser2024modelcollapseinevitablebreaking,
      title={Is Model Collapse Inevitable? Breaking the Curse of Recursion by Accumulating Real and Synthetic Data}, 
      author={Matthias Gerstgrasser and Rylan Schaeffer and Apratim Dey and Rafael Rafailov and Henry Sleight and John Hughes and Tomasz Korbak and Rajashree Agrawal and Dhruv Pai and Andrey Gromov and Daniel A. Roberts and Diyi Yang and David L. Donoho and Sanmi Koyejo},
      year={2024},
      eprint={2404.01413},
      archivePrefix={arXiv},
      primaryClass={cs.LG},
      url={https://arxiv.org/abs/2404.01413}, 
}

\newpage
\appendix

\section{Prompt for LLM-as-a-judge in quality filtering}\label{app:llm-judge-prompt}

\begin{promptbox}[Educational filter -- LLM-as-a-judge prompt (1/2)]

SYSTEM: Deine Aufgabe ist es zu bewerten, wie gut die deutsche Sprachqualität eines deutschen Textauszugs ist.

Gebe deine Bewertung in folgendem JSON-Format: \\
\{ \\
    \hspace*{1.5em}"content\_criticism": str (Falls es Schwächen der Antwort im Bezug auf Inhalt gibt, zum Beispiel plötzlicher Themenwechsel, unklarer Sinn, oder fehlende Informationen, nenne und belege sie anhand von Beispielen.), \\
    
    \hspace*{1.5em}"content\_grade": number (Eine Zahl zwischen 1 und 5, wobei 5 die beste und 1 die schlechteste Bewertung ist: \\
    \hspace*{1.5em}5: Sehr kohärent und informativ, sehr hohe Qualität. \\
    \hspace*{1.5em}4: Kohärent und informativ. \\
    \hspace*{1.5em}3: Kleine Schächen im Inhalt, aber der Sinn kann erfasst werden. \\
    \hspace*{1.5em}2: Begrenzte Relevanz oder Genauigkeit, der Sinn kann nicht ganz erfasst werden. \\
    \hspace*{1.5em}1: Völlig inkonsistent oder unsinnig. \\
), \\
\{ \\
    \hspace*{1.5em}"language\_criticism": str (Falls es Schwächen der Antwort im Bezug auf Sprache gibt, zum Beispiel Mischung verschiedener Sprachen, Schimpfwörter, unübliche Wortwahl, nenne und belege sie anhand von Beispielen.), \\
    
    \hspace*{1.5em}"language\_grade": number (Eine Zahl zwischen 1 und 5, wobei 5 die beste und 1 die schlechteste Bewertung ist: \\
    \hspace*{1.5em}5: Sehr förmliche, objektive deutsche Sprache. \\
    \hspace*{1.5em}4: Wenige, vernachlässigbare Sprachfehler.
\end{promptbox}
\begin{promptbox}[Educational filter -- LLM-as-a-judge prompt (2/2) continued]
    \hspace*{1.5em}3: Es gibt sprachliche Schwächen, aber der Sinn kann gut 
    erfasst werden. \\
    \hspace*{1.5em}2: Stark umgangssprachlich/Sprachmischung. \\
    \hspace*{1.5em}1: Grobe Sprachmischung, Schimpfwörter oder schlechte Wortwahl, der Sinn kann nicht ganz erfasst werden. \\
), \\
    \hspace*{1.5em}"orthography\_criticism": str (Falls es Schwächen der Antwort im Bezug auf Grammatik und Rechtschreibung oder Tippfehler gibt, nenne und belege sie anhand von Beispielen. Dazu gehören Worte, die getrennt geschrieben wurden obwohl sie zusammengeschrieben werden sollten, falsche Verbbeugung und falsche Deklination, etc.), \\
    
    \hspace*{1.5em}"orthography\_grade": number (Eine Zahl zwischen 1 und 5, wobei 5 die beste und 1 die schlechteste Bewertung ist: \\
    \hspace*{1.5em}5: Keine Fehler. \\
    \hspace*{1.5em}4: wenige unbedeutende Fehler. \\
    \hspace*{1.5em}3: einige unbedeutende Fehler. \\
    \hspace*{1.5em}2: ein paar grobe Fehler, aber die Antwort ist verständlich. \\
    \hspace*{1.5em}1: so viele Fehler, dass die Antwort unverständlich ist. \\
), \\

USER: Textauszug: \\
\{document\}

ASSISTANT: Bewertungs-JSON:

\{
\end{promptbox}

\section{Prompts for synthetic data generation}\label{app:synth-gen-prompt}

\begin{promptbox}[Rephrasing -- data generation prompt] % to_prompt_0
Geben Sie mir für den folgenden Absatz eine vielfältige Umformulierung in hochwertiger deutscher Sprache, im Stil von Wikipedia.
Beginne deine Antwort mit 'Umformulierung:'.

Text:

\{document\}

Umformulierung:
\end{promptbox}

\begin{promptbox}[Summarisation -- data generation prompt] % to_prompt_2
Ihre Aufgabe ist es, den bereitgestellten Text gemäß diesen Anweisungen zu lesen und umzuformulieren:

- Versuchen Sie, eine komprimierte, aber genaue und informative Version des Originaltextes zu erstellen, keine vereinfachte Zusammenfassung.

- Erfassen und bewahren Sie die entscheidenden Informationen, Schlüsselkonzepte, wichtigen Werte und sachlichen Details im Originaltext auf und machen Sie ihn gleichzeitig lesbarer und zugänglicher.

- Behalten Sie Fachbegriffe, Fachvokabular und komplexe Konzepte bei.

- Bewahren Sie Beispiele, Erklärungen zu Denkschritten und unterstützende Beweise auf, um die Tiefe und den Kontext des Textes zu erhalten.

- Fügen Sie nur Informationen hinzu, die im Originaltext vorhanden sind. Fügen Sie keine neuen oder unbegründeten Behauptungen hinzu.

- Schreiben Sie im Klartext

Beginne deine Antwort mit 'Umformulierte Version:'.

Text:

\{document\}

Umformulierte Version:

\end{promptbox}

\begin{promptbox}[Rephrasing in Wikipedia style -- data generation prompt] % to_prompt_4
Ihre Aufgabe ist es, einen neuen Text zu Wissen aus dem bereitgestellten Text zu verfassen, indem Sie diesen Anweisungen folgen:

- Schreibe den Text als Passagen mit leicht verständlichen und qualitativ hochwertigen deutsche Sätzen um, wie sie in Lehrbüchern und Wikipedia stehen.

- Konzentrieren Sie sich auf inhaltliche Disziplinen wie Geisteswissenschaften, Sozialwissenschaften, Naturwissenschaften, Technik, Ingenieurwesen, Mathematik, Recht und Recht, Wirtschaft, Management, Kunst, Bildung, Agrarwissenschaften, Politik und Geschichte.

- Ignorieren Sie Inhalte, die keine nützlichen Fakten oder Kenntnisse enthalten.

- Bewahren Sie Beispiele, Erklärungen von Denkprozessen und unterstützende Beweise auf, um die Tiefe und den Kontext des Textes zu erhalten.

- Fügen Sie keine Details hinzu oder ändern Sie sie. Wiederholen Sie nur, was bereits im Text steht.

- Schreiben Sie im Klartext.

- Fügen Sie keine Titel, Untertitel, Notizen oder Kommentare hinzu.

Nacheinander wird der Nutzer nun Texte bereitstellen, verfolge dann die Anweisungen und schreibe den neuen Text.

Beginne deine Antwort mit 'Neuer Text:'.

Text:

\{document\}

Neuer Text:
\end{promptbox}

\begin{promptbox}[Formulating questions -- data generation prompt] % to_prompt_1
Lesen Sie den bereitgestellten Text, und stellen Sie allgemeine Fragen. Befolgen Sie die folgenden Anweisungen sorgfältig:

1. Die Fragen sollen auch ohne den Text als Allgemeinwissen beantwortet werden können.

2. Wenn das nicht möglich ist, formuliere zunächst einen kurzen Kontext, oder antworte lediglich 'nicht möglich'.

3. Stellen Sie möglichst diverse Fragen zu unterschiedlichen sachlichen Informationen, wichtigem Wissen oder konkreten Details im Text.

4. Beantworten Sie die Fragen direkt und in kurzer prägnanter Sprache.

5. Stellen Sie Fragen in den folgenden verschiedenen Kategorien: Ja/Nein, Multiple-Choice mit mehreren Optionen zur Auswahl, Vergleichs- und Offener Fragen.

6. Stellen Sie 8 Fragen mit Antworten, zwei zu jeder der beschriebenen Kategorien.

7. Beginne mit 'Fragen-Antwort-Paare:'.

Text:

\{document\}

Fragen-Antwort-Paare:
\end{promptbox}

\begin{promptbox}[Extracting lists -- data generation prompt] % to_prompt_3
Überprüfen Sie den Text, und extrahieren Sie die wichtigsten Informationen auf Deutsch. Befolgen Sie diese Anweisungen:

- Lesen Sie den obigen Text sorgfältig durch und erstellen Sie eine prägnante und organisierte Liste mit sachlichen Informationen, konkreten Details, Schlüsselkonzepten sowie wichtigen Zahlen und Statistiken, die aus dem Text entnommen sind.

- Stellen Sie sicher, dass jeder Punkt klar und spezifisch ist und durch den Originaltext gestützt wird.

- Stellen Sie sicher, dass der extrahierte Text informationsdicht und leichter zu erlernen ist.

- Fügen Sie keine Titel oder Überschriften hinzu.
Beginne deine Antwort mit 'Liste:'.

Text:

\{document\}

Liste:

\end{promptbox}

\section{Experiments}\label{app:experiment-configs}

\subsection{1B Llama-like model training details}\label{app:1B-training-details}

We trained a 1 billion parameter transformer model with 16 layers, 2,048 hidden dimensions, 32 attention heads, and 8 key-value heads. The model uses a context length of 4096 tokens and employs modern architecture choices: rotary positional embeddings (RoPE) with a base of 500,000, SwiGLU activation functions with an MLP expansion factor of 4, and RMSNorm for layer normalisation.

The model was trained for 40,000 iterations with a global batch size of 512. We used the AdamW optimiser with $\beta_1=0.9$, $\beta_2=0.95$, $\epsilon=1e-8$, and gradient clipping at 1.0. The learning rate followed a cosine decay schedule with an initial rate of 3e-4, 400 warmup steps, 4,000 cooldown steps, and minimum learning rate of 0.0. We implemented ZeRO optimisation for distributed training efficiency.
The model uses BFloat16 precision and was trained using a UTF-8 tokeniser with a vocabulary size of 131,072 tokens.

We used the Pharia-1 tokeniser that was trained with the Unigram algorithm using the SentencePiece library \cite{alephalpha_pharia1_2024}.

\begin{table}[h]
\renewcommand{\arraystretch}{1.5}
\centering
\begin{tabular}{lc}
\hline
\textbf{Parameter} & \textbf{Value} \\ \hline
$\beta_1$ & 0.9 \\
$\beta_2$ & 0.95 \\
$\epsilon$ & 1e-8 \\
$N_{\text{Warmup}}$ & 1\% \\
$N_{\text{Cooldown}}$ & 10\% \\
learning rate & 3e-4 \\ \hline
\end{tabular}
\caption{Parameters of the AdamW optimiser for the 1B-Llama-model.}
\end{table}

\begin{table}[h]
\renewcommand{\arraystretch}{1.5}
\centering
\begin{tabular}{lc}
\hline
\textbf{Hyperparameter} & \textbf{Value} \\ \hline
$N_\text{vocab}$ & 131,072 \\
$n_{\text{Layers}}$ & 16 \\
$n_{\text{heads}}$ & 32 \\
$d_{\text{model}}$ & 2,048 \\
$d_{\text{MLP}}$ & 8,192 \\
Batch Size & 512 \\
Sequence Length & 4,096 \\ \hline
\end{tabular}
\caption{Hyperparameters for the 1B-Llama-model.}
\end{table}

\subsection{8B HAT model training details}\label{app:8B-training-details}

For the second part of our experiments, we trained 8B HAT models that consist of 3 consecutive transformers. In contrast to classical tokeniser-based models, HAT splits words by a rule-based splitter. Words are processed byte-wise by the smaller outer transformers, while the bigger backbone acts as the next-word predictor on a single embedding per word.
The first and third models act as encoder and decoder, that transform bytes to word embeddings and vice versa. These models are connected by linear projection layers.

Words are split by special characters and punctuation, and converted into UTF-8 byte sequences.

All three models are trained simultaneously end-to-end,
with hyperparameters similar to those in §\ref{app:8B-training-details} and shown in the following tables.

\begin{table}[h]
\renewcommand{\arraystretch}{1.5}
\centering
\begin{tabular}{lc}
\hline
\textbf{Parameter} & \textbf{Value} \\ \hline
$\beta_1$ & 0.9 \\
$\beta_2$ & 0.95 \\
$\epsilon$ & 1e-8 \\
$N_{\text{Warmup}}$ & 1\% \\
$N_{\text{Cooldown}}$ & 10\% \\
learning rate & 1e-4 \\ \hline
\end{tabular}
\caption{Shared parameters of the AdamW optimiser for all transformer modules assembling the 8B-HAT-model.}
\end{table}

\begin{table}[h]
\renewcommand{\arraystretch}{1.5}
\centering
\begin{tabular}{lc}
\hline
\textbf{Hyperparameter} & \textbf{Value} \\ \hline
$N_\text{vocab}$ & 256 \\
$n_{\text{Layers}}^{\text{Encoder}}$ & 6 \\
$n_{\text{Layers}}^{\text{Decoder}}$ & 4 \\
$n_{\text{heads}}$ & 6 \\
$d_{\text{model}}$ & 768 \\
$d_{\text{MLP}}$ & 2,047 \\
Batch Size & 1,024 \\
Sequence Length & \textit{variable} \\ \hline
\end{tabular}
\caption{Hyperparameters for the Encoder/ Decoder Models of 8B-HAT. Almost all parameters are the same for the Encoder and Decoder. Only the number of layers is increased to 6 for the Encoder. As we fixate the number of words per training steps in the Backbone, the Encoder and Decoder will see a variable sequence length per training batch.}
\end{table}

\begin{table}[h]
\renewcommand{\arraystretch}{1.5}
\centering
\begin{tabular}{lc}
\hline
\textbf{Hyperparameter} & \textbf{Value} \\ \hline
$N_\text{vocab}$ & - \\
$n_{\text{Layers}}$ & 32 \\
$n_{\text{heads}}$ & 32 \\
$d_{\text{model}}$ & 4,096 \\
$d_{\text{MLP}}$ & 11,008 \\
Batch Size & 1,024 \\
Sequence Length & 3,500 \\ \hline
\end{tabular}
\caption{Hyperparameters for the Backbone Model of 8B-HAT.}
\end{table}

\subsection{HQ datasets list}\label{app:HQ-datasets-list}

Books \& literature (\eg, Gutenberg); legal \& government
(\eg, European Patent Office (EPO) and EU Parliament); news \& web text (\eg, Wikinews); encyclopedic \& reference
(\eg, Wikipedia); scientific \& technical (\eg, peS2o); and parallel \& multilingual (\eg, ParaCrawl).

\section{Extended results}

\subsection{1B Llama-style model individual benchmarks results}\label{app:1B-benchmarks}

\dataset{} outperforms FineWeb2 across all benchmarks. Interestingly, we find that different subsets have different strengths: our synthetic data performed particularly well on MMMLU, whereas our filtered CC and high-quality classified data performed best for HellaSwag and ARC.

\begin{figure*}[h]
    \centering
    \includegraphics[width=1\linewidth]{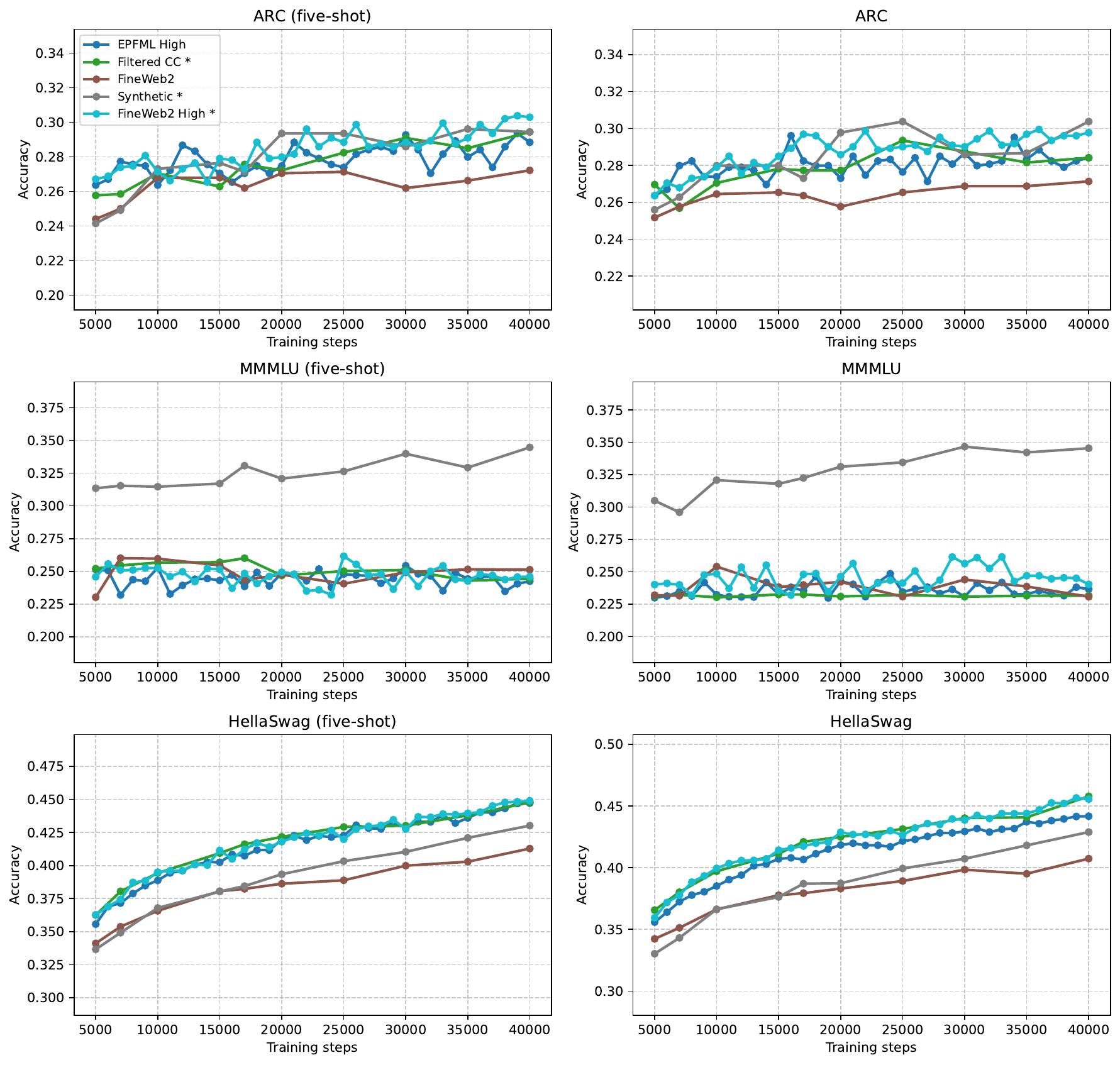}
    \caption{Individual benchmark results for 1B Llama-style models trained on $\sim$ 84 billion tokens from different datasets. Here, 1,000 training steps equates to approximately 2.1 billion tokens.}
    \label{fig:German-individual-benchmarks}
\end{figure*}

\subsection{Compute Resources}\label{apx:compute}

The training of our 1B-model for 40k steps, which in total amounts to roughly $8.4\times10^{10}$ tokens, was performed for 9 datasets: Our 5 FineWeb2 quality buckets, random Fineweb, epfml-FineWeb2, Filtered CC, and our synthetic data. Computation was done with current standards in bfloat16 format. The total compute budget was 12 hours on 64 H100 GPUs per ablation.

The 8B HAT-model was trained for 3 different experiments with 5 runs in total. First, for 50k steps, which in total amount to roughly $1.5\times10^{11}$ words for Filtered CC and FineWeb2. Secondly, for 21k steps, around $6.3\times10^{10}$ words for synthetic data. The 21k snapshot from the FineWeb2 training was reused for this comparison. Lastly for 25k on English and 20k on German data, equating to roughly $1.35\times10^{11}$ words in total. Note that each word is one token. The total compute budget was 4 days on 128 H100 GPUs per ablation.

\end{document}